\documentclass[10pt,twocolumn,letterpaper]{article}

\usepackage[pagenumbers]{iccv} % To force page numbers, e.g. for an arXiv version

\definecolor{iccvblue}{rgb}{0.21,0.49,0.74}
\usepackage[pagebackref,breaklinks,colorlinks,allcolors=iccvblue]{hyperref}

\usepackage{graphicx}
\usepackage{colortbl}
\usepackage{subcaption}
\usepackage{booktabs}
\usepackage{multirow}
\usepackage{makecell}
\usepackage{hhline}
\usepackage{hyperref}

\title{FastInstShadow: A Simple Query-Based Model for Instance Shadow Detection}

\author{Takeru Inoue$^{1}$ \quad Ryusuke Miyamoto$^{2}$\\
$^{1}$Department of Computer Science, Graduate School of Science and Technology, Meiji University, Japan\\
$^{2}$Department of Computer Science, School of Science and Technology Meiji University, Japan \\
{\tt\small \{takeru, miya\}@cs.meiji.ac.jp}
}

\def\proc{Proc. }

\def\icip#1{\proc ICIP}

\begin{document}
\maketitle
\begin{abstract}
  Instance shadow detection is the task of detecting pairs of shadows and objects,
  where existing methods first detect shadows and objects independently, then associate them.
  This paper introduces FastInstShadow, a method that enhances detection accuracy through a query-based architecture
  featuring an association transformer decoder with two dual-path transformer decoders to assess relationships between shadows and objects during detection. 
  Experimental results using the SOBA dataset showed that the proposed
  method outperforms all existing methods across all criteria.
  This method makes real-time processing feasible for
  moderate-resolution images with better accuracy
  than SSISv2, the most accurate existing method.
  Our code is available at \small{\url{https://github.com/wlotkr/FastInstShadow}}.
\end{abstract}

\section{Introduction}
Instance shadow detection methods detect shadows and their corresponding objects as pixel-wise class labels and identifies pairs between them.
This is an essential process for various applications.
For example, in photo editing, shadow detection allows manipulation of
unified objects and their associated shadows, improving flexibility in editing operations such as deletion, scaling,
and translation \cite{bib:lisa, bib:ssisv2, bib:vishadow}.
Some image generation tasks, including shadow removal, shadow generation, and object compositing, require assistance from instance shadow detection methods.
A majority of shadow removal~\cite{bib:shadowdiffusion, bib:shadowformer} methods require the shadow masks as input,
which can be easily obtained through instance shadow detection methods.
Shadow generation~\cite{bib:sgdiffusion, bib:dmasnet, bib:sgrnet} also needs the assistance from instance shadow detection methods
because their accuracy can be improved by using binary masks of background objects and associated shadows as clues.
Instance shadow detection methods are also effectively utilized in both the creation of
the DESOBAv2~\cite{bib:sgdiffusion} dataset for shadow generation,
and datasets for realistic object placement with shadow effects~\cite{bib:object-compositing}.
This emphasizes the importance of associating shadows with corresponding objects.
Therefore, advances in instance shadow detection will also contribute to several types of vision tasks such as image generation,
one of the hottest tasks since the rise of diffusion models~\cite{bib:diffusion-model, bib:ddpm}.

\begin{figure}[t]
  \centering
    \includegraphics[width=0.90\linewidth]{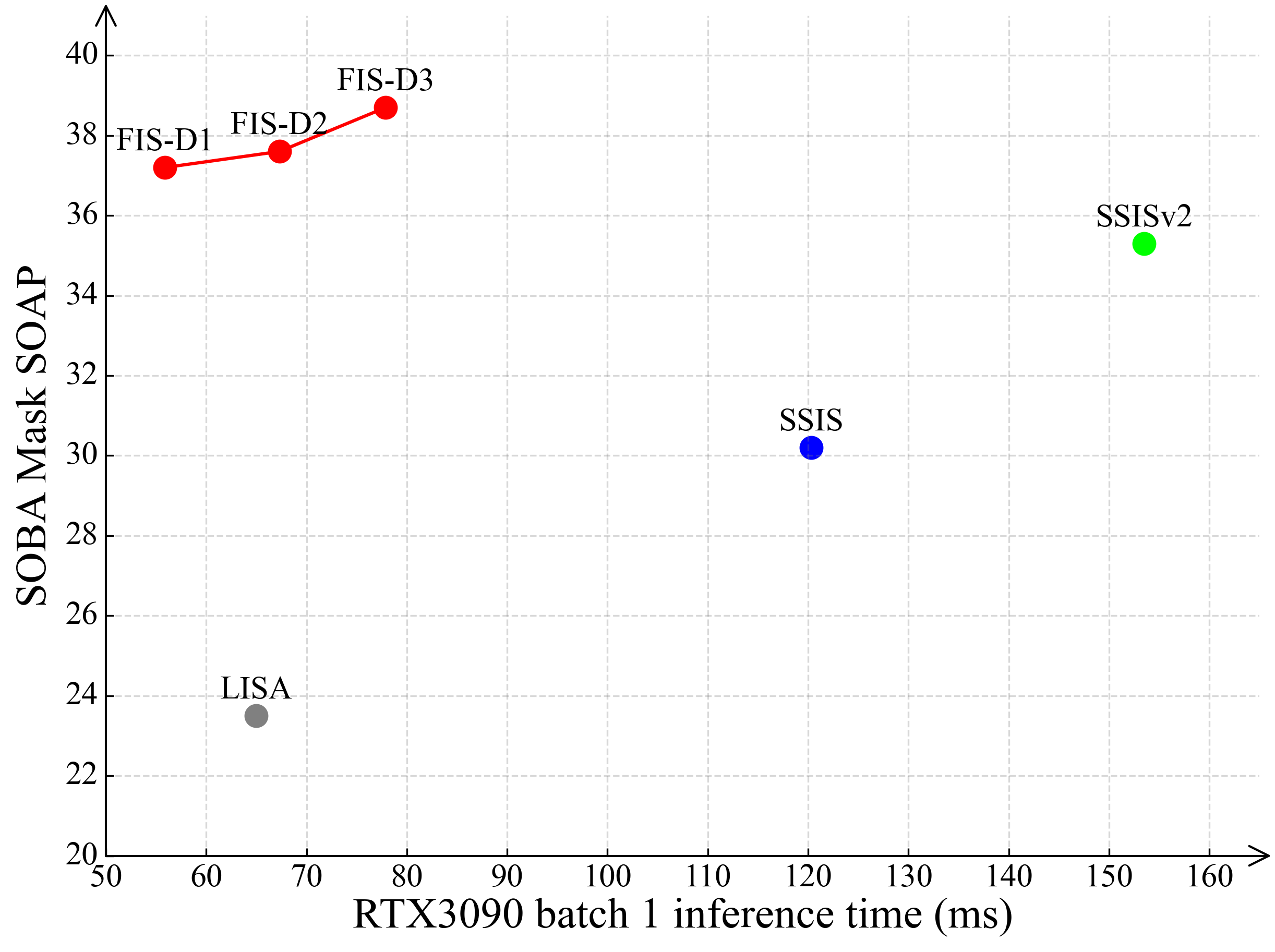}
    \caption{\textbf{Speed-performance trade-off on SOBA-testing.}
    Our FastInstShadow (FIS) is a scalable framework where the lightweight variant,
    FIS-D1, outperforms all existing methods in terms of both speed and accuracy on $SOAP_{segm}$~\cite{bib:lisa}.
    The larger-scale variants, FIS-D2 and FIS-D3, achieve further accuracy improvements.}
    \label{fig:speed-performance}
\end{figure}

Existing instance shadow detection methods \cite{bib:lisa, bib:ssis, bib:ssisv2}
attempt to solve this task through a two-stage approach:
individual pixel-wise detection of shadows and objects,
and pairing detected shadows with their corresponding detected objects.
These methods extend instance segmentation models by incorporating pairing processes:
detection is performed by an instance segmentation model, and pairing is done by an incorporated process.

Considering the current approaches to instance shadow detection,
the detection and pairing accuracy strongly depends on the instance segmentation model and extended pairing process, respectively.
However, despite recent attention in instance segmentation focusing on query-based methods
due to their non-maximum suppression (NMS)-free simple pipeline and high detection accuracy,
no methods have yet incorporated query-based models into instance shadow detection methods.
Additionally, treating detection and pairing as separate processes complicates the inference pipeline
and potentially misses opportunities to learn mutual relationships of shadows and objects.

To overcome issues with existing instance shadow detection methods,
this paper proposes a novel method named FastInstShadow (FIS) with sufficient accuracy and processing speed for practical applications.
FIS is based on FastInst~\cite{bib:fastinst}, a query-based instance segmentation method
that delivers real-time performance and high accuracy through a simple architecture.
Our key innovation is designing queries that directly aggregate
shadow and object features, enabling the model to learn their mutual
relationships, detecting paired shadow-object instances.
The query design makes it possible to remove the pairing process
of shadows and objects that is indispensable for existing methods.
Additionally, we incorporate two training strategies that improve accuracy
without adding complexity the inference pipeline.
As shown in \cref{fig:speed-performance}, our method achieves strong
performance on the SOBA-testing~\cite{bib:lisa}.

Our main contributions can be summarized as follows:
\begin{itemize}
  \item We propose the first query-based model for instance shadow detection
    having an association transformer decoder composed of two dual-path
    transformer decoders~\cite{bib:fastinst}, eliminating the pairing processes that have been indispensable for instance shadow detection methods.
  \item We designed shadow direction learning and box-aware mask loss processes to improve the accuracy of instance shadow detection.
  \item Despite its simple architecture, our method is significantly more accurate than state-of-the-art(SOTA) methods.
  \item Inference speed is drastically improved, reaching more than
    30 frames per second(fps) for images with moderate resolutions,
    with higher accuracy than SOTA methods.
\end{itemize}

\section{Related Work}
This section explains the taxonomy of instance segmentation and
instance shadow detection, which is closely related to
the contribution of the proposed method.

\subsection{Instance segmentation}
Existing instance shadow detection methods are based on instance segmentation models,
which are vital in detecting shadows and objects.
Current instance segmentation models can be grouped into
region-based methods, instance activation-based methods, and query-based methods.

\textbf{Region-based methods} first detect bounding boxes for instances
and then extract region features using RoI-Pooling~\cite{bib:faster-rcnn} or RoI-Align~\cite{bib:mask-rcnn} for pixel-wise detection
\cite{bib:panet, bib:htc, bib:cascade-rcnn, bib:pointrend, bib:mask-scoring}.
As a pioneering work, Mask R-CNN~\cite{bib:mask-rcnn} extends Faster R-CNN~\cite{bib:faster-rcnn} by adding a mask branch for pixel-wise detection.

\textbf{Instance activation-based methods} represent instances by semantically meaningful pixels and perform pixel-wise detection using the features of these pixels
\cite{bib:solo, bib:solov2, bib:meinst, bib:tensormask, bib:sparseinst, bib:centermask, bib:yoloact-plusplus}.
For example, CondInst~\cite{bib:condinst} extends FCOS~\cite{bib:fcos} by predicting mask embedding vectors for dynamic convolution for pixel-wise detection.

\textbf{Query-based methods} represent instances as queries and perform pixel-wise detection through transformer decoders
\cite{bib:solq, bib:istr, bib:mask2former, bib:maskdino, bib:segformer}.
Recent attention in instance segmentation has focused on query-based methods due to their simple, NMS-free pipelines and high accuracy.
While early query-based methods were limited by slow inference speeds,
FastInst~\cite{bib:fastinst} achieves real-time performance through architectural refinements.

\subsection{Instance shadow detection}
We present details of both existing instance shadow detection stages:
the first utilizes instance segmentation and the second implements the pairing process.

\textbf{Region-based methods} were first introduced to instance shadow detection through LISA~\cite{bib:lisa},
an extension of Mask R-CNN~\cite{bib:mask-rcnn} featuring a separate branch for detecting integrated regions for each shadow-object pair and a heuristic pairing algorithm.
The main branch inherited from Mask R-CNN detects individual shadows and objects, then a heuristic pairing algorithm associates them.
The pairing algorithm selects candidate pairs when the distance between detected shadows and objects is below an empirical threshold.
The algorithm finalizes pairs by evaluating the IoU scores between the merged regions of candidate pairs and the detections from the separate branch.

\textbf{Instance activation-based methods} were introduced to instance shadow detection through SSIS\cite{bib:ssis} and SSISv2\cite{bib:ssisv2}.
These methods extend CondInst with a separate head for detecting associated instances and a heuristic pairing algorithm.
The main head inherited from CondInst detects individual shadows and objects, while the separate head detects instances paired with the main head's instances.
Thus, these methods detect shadows and objects as pairs; however, the detection accuracy of the separate head is lower than that of the main head.
Therefore, these methods also employ a heuristic pairing algorithm to pair shadow and object detections from the main head.
However, the relationships between shadow and objects detected by the main head are not considered in the detection stage.

Region-based and instance activation-based methods represent
instances using bounding boxes and semantically meaningful pixels, respectively.
As, these methods cannot address shadow and object dependencies,
we introduce a query-based instance shadow detection method
that has gained popularity in the field of instance segmentation
which can treat relationships between shadows and objects appropriately.

\subsection{FastInst Architecture}
FastInst~\cite{bib:fastinst} follows the meta-architecture of Mask2Former~\cite{bib:mask2former} and integrates three key innovations:\\
(i) \textbf{Instance activation-guided queries} dynamically select pixel embeddings with high semantics from feature maps as initial queries, enriching embedding information.
This reduces the iteration update burden of the transformer decoder.\\
(ii) \textbf{Dual-path update strategy} alternately updates query and pixel features, enhancing the representation ability of pixel features.
This saves us from the heavy pixel decoder design and reduces dependence on the number of decoder layers.\\
(iii) \textbf{Ground truth mask-guided learning} replaces the mask used in the standard masked attention with the last layer bipartite matched ground truth mask.
This allows each query to see the whole region of its predicted target instances during training.

These innovations enable FastInst to achieve high accuracy and real-time instance segmentation.
Similarly, we leverage these innovations in our method to achieve high accuracy and real-time instance shadow detection.

\section{Methods}
In this section, we present our FIS method, based on FastInst.
We describe two strategies employed to train this architecture:
shadow direction learning and box-aware mask loss.

\begin{figure*}[t]
    \centering
    \includegraphics[width=\linewidth]{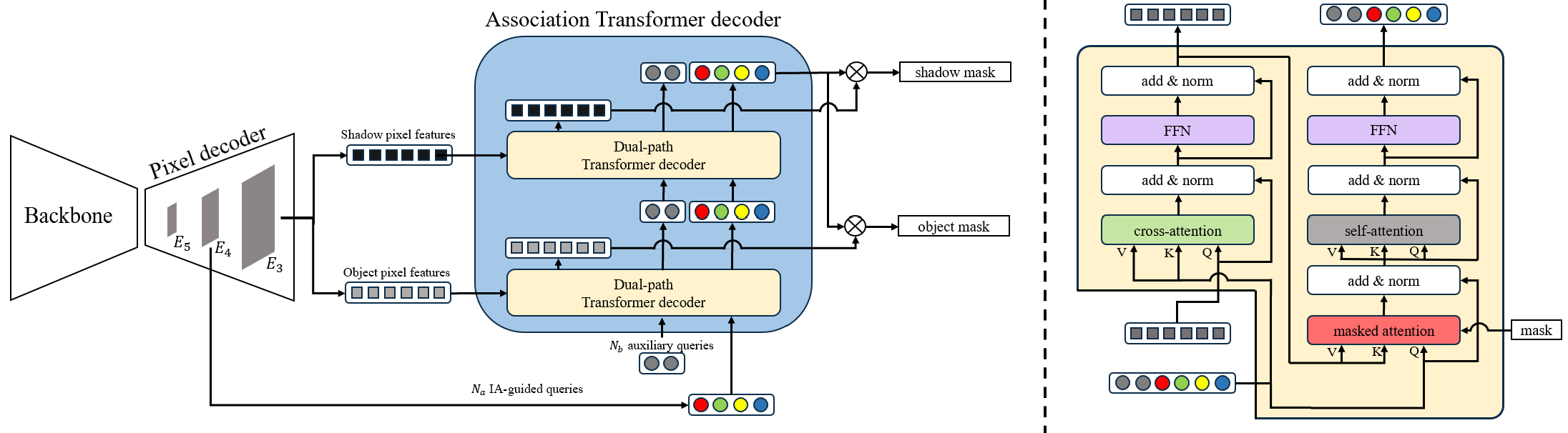}
    \caption{\textbf{FastInstShadow(FIS) architecture.}
    The left side of the figure illustrates the overall architecture, while the right side details the dual-path transformer decoder~\cite{bib:fastinst}.
    FIS consists of three components: a backbone network and a pixel decoder, both inherited from FastInst~\cite{bib:fastinst},
    along with our novel association transformer decoder.
    Furthermore, object and shadow pixel features, obtained by flattening the feature map $E_3$,
    along with queries designed similar to FastInst, are input into the association transformer decoder.
    The association transformer decoder consists of two dual-path Transformer decoders:
    the first processes object pixel features to capture object characteristics and
    the second processes shadow pixel features to capture shadow characteristics.
    This design enables the model to detect shadows and objects as paired instances while considering their mutual relationships.
    }
    \label{fig:fis}
\end{figure*}

\subsection{Overall architecture}
\Cref{fig:fis} illustrates the architecture of FIS.
Our model first extracts three feature maps $C_3$, $C_4$ and $C_5$, where resolutions are $1/8$, $1/16$ and $1/32$
of the input image $I \in \mathbb{R}^{H \times W \times 3}$, respectively,
using ResNet-50-d~\cite{bib:bag-of-tricks} with deformable convolutions~\cite{bib:deformable-conv} as a backbone.
Three feature maps are projected to $256$ channel maps by a $1 \times 1$ convolutional layer.
PPM-FPN~\cite{bib:sparseinst}, as a pixel decoder, accepts these feature maps as input and then outputs multi-scale feature maps $E_3$, $E_4$ and $E_5$.
Subsequently, $N_a$ instance activation-guided queries~\cite{bib:fastinst} select pixel embeddings with higher object semantics.
These queries are concatenated with $N_b$ auxiliary learnable queries
to obtain the total queries $Q \in \mathbb{R}^{N \times 256}$, where $N = N_a + N_b$. 
Additionally, the high-resolution pixel feature $E_3$ is flattened to produce
object pixel features $X_o \in \mathbb{R}^{L \times 256}$ and shadow pixel features $X_s \in \mathbb{R}^{L \times 256}$, where $L = H/8 \times W/8$.
The association transformer decoder then accepts these features as input for performing pixel-wise detection of shadows and objects as pairs.

\subsection{Association transformer decoder}
The association transformer decoder consists of two dual-path transformer decoders~\cite{bib:fastinst} connected sequentially.
The two dual-path transformer decoders focus on object and shadow features, respectively.
When queries $Q$, object pixel features $X_o$, and shadow pixel features $X_s$ are input to the association transformer decoder,
positional embeddings are added to these features.
Next, queries $Q$ and object pixel features $X_o$ are input to the first dual-path transformer decoder.
Here, object pixel features $X_o$ are updated through the cross-attention layer on queries $Q$ and the feedforward layer.
Then, queries $Q$ are updated through masked attention, self-attention, and the feedforward layer.
Masked attention aggregates object features into queries $Q$ using predicted object masks from the previous layer.

Furthermore, updated queries $Q$ and shadow pixel features $X_s$ are input to the second dual-path transformer decoder.
Similar to the first decoder, shadow pixel features $X_s$ are updated through the cross-attention layer on updated queries $Q$ and the feedforward layer.
At this stage, queries $Q$ already contain object features, assisting the transformer decoder in detecting shadows that correspond to the objects.
Then, similar to the first decoder, queries $Q$ are updated through masked attention, self-attention and feedforward layer.
Masked attention aggregates shadow features into queries $Q$ using predicted shadow masks from the previous layer.
The association transformer decoder takes object pixel features $X_o$, shadow pixel features $X_s$, and queries $Q$ as input
and outputs updated object pixel features $X^\prime$, shadow pixel features $X_s^\prime$, and queries $Q^\prime$.
These updated features are then used as input to the next Association Transformer decoder.
By repeatedly employing the association transformer decoder, queries $Q$ aggregate object and shadow features, capturing their mutual relationships.

During inference, a three-layer MLP is applied to refined queries $Q$ to obtain mask embeddings.
A linear projection is applied to refined object pixel features $X_o$ and shadow pixel features $X_s$
to obtain object and shadow mask features, respectively.
Then, object and shadow masks for each query are obtained by multiplying mask embeddings with each mask feature.

\subsection{Training strategies}
\subsubsection{Shadow direction learning}
In the FIS architecture, the association transformer decoder learns the mutual relationships between shadows and objects.
To further facilitate relation learning, we implement a directional learning strategy from objects to shadows.
In this strategy, an offset vector map is first generated by reducing the channel dimension of the output from the pixel decoder $E_3$ through a $3 \times 3$ convolutional layer.
This offset vector map contains two channels representing $x$ and $y$ coordinates.
The offset vector map is denoted as $V \in \mathbb{R}^{H' \times W' \times 2}$, where $H' = H / 8$ and $W' = W / 8$.
Subsequently, center coordinates of object masks obtained by the association transformer decoder are computed
and the corresponding offset vectors are extracted from the offset vector map $V$ at these coordinates.
The model is then trained to match the sum of each query's object mask center coordinates and offset vector with the shadow mask's ground-truth center coordinates.

Our proposed shadow direction loss $\mathcal{L}_\text{d}$ is defined as:
\begin{equation}
    \begin{aligned}
        \mathcal{L}_\text{d} =
        \begin{cases} 
        \lambda_\text{d}(0.5 (\theta^p - \theta^g)^2 / \beta), & \text{if } |\theta^p - \theta^g| < \beta \\
        \lambda_\text{d}(|\theta^p - \theta^g| - 0.5 \cdot \beta), & \text{otherwise},
        \end{cases}
    \end{aligned}
    \label{equ:shadow-direction-loss}
\end{equation}
where $\lambda_\text{d}$ is the loss weight, $\beta$ is a hyperparameter,
and $\theta^p$ and $\theta^g$ are the predicted and ground-truth coordinates of the shadow, respectively.
The smooth $\mathcal{L}_{1}$ loss~\cite{bib:fast-rcnn} is used as shadow direction loss.

This strategy for learning bidirectional relationships between shadows and objects has also been employed
in LISA~\cite{bib:lisa}, SSIS~\cite{bib:ssis}, and SSISv2~\cite{bib:ssisv2}.
However, while SSIS and SSISv2 transform inferred coordinates into a map and use it as input to the mask head,
our method does not utilize inferred coordinates for shadow and object detection, similar to LISA.
This approach allows us to maintain a simple pipeline during inference while improving accuracy.

\subsubsection{Box-aware mask loss}
Our architecture does not include a mechanism for predicting bounding boxes of shadows and objects,
unlike existing instance shadow methods~\cite{bib:lisa, bib:ssis, bib:ssisv2}.
Consequently, generated masks may not consider their bounding boxes,
leading to false positive detections that deviate significantly from ground truth regions.
In particular, shadow detection is challenging
due to similar shadow textures regardless of which objects cast them, as well as irregular shapes caused by lighting conditions,
making these masks prone to such deviant false positive detections.
Thus, we introduce a reweighting strategy in binary mask cross-entropy loss.

Our proposed loss function is defined as:
\begin{equation}
    \begin{aligned}
        \mathcal{L}_{\text{ce}} &= -\frac{1}{K} \sum_{i=1}^{K} w_{i} \left[y_{i} \cdot \log(p_{i}) + (1 - y_{i}) \cdot \log(1 - p_{i})\right] \\
        w_{i} &= \begin{cases}
            \alpha & 
            \begin{aligned}
                & \text{if point $i$ is outside the AABB} \\ 
                & \text{and $p_{i} > 0.5$ where $y_{i} = 0$}
            \end{aligned} \\
            1 & \text{otherwise}
        \end{cases}
    \end{aligned}
    \label{equ:suppress-false-positive-loss}
\end{equation}
In our method, similar to \cite{bib:fastinst, bib:mask2former}, the mask loss is computed using $K$ randomly sampled points.
$y_i$, $p_i$ and $w_i$ represent the target label, the predicted probability and the weight at the sampled point $i$, respectively.
This loss function assigns a larger value to $w_{i}$
when point $i$ is outside the axis-aligned bounding box (AABB) of the corresponding ground truth mask
and incorrectly predicted as positive.
This weighting scheme penalizes deviant false positive pixels, thereby suppressing such detections.
However, applying this weighting scheme in early training phase can result in inaccurate training.
Therefore, we only apply this loss function after $N$ epochs of training, using the conventional loss function with $w_i = 1$ until epoch $N$.

\begin{table*}[t]
    \centering
    \begin{tabular}{l|c|cc|cc|cc}
        \hline
        Network & fps & $SOAP_{segm}$ & $SOAP_{bbox}$ & Assoc. $AP_{segm}$ & Assoc. $AP_{bbox}$ & Inst. $AP_{segm}$ & Inst. $AP_{bbox}$ \\
        \hline
        LISA~\cite{bib:lisa}   & 15.38 & 23.5 & 21.9 & 40.9 & 48.4 & 39.2 & 37.6 \\
        SSIS~\cite{bib:ssis}   & 8.31  & 30.2 & 27.1 & 52.2 & 59.6 & 43.4 & 41.3 \\
        SSISv2~\cite{bib:ssisv2} & 6.51  & 35.3 & 29.0 & 59.2 & 63.1 & 50.2 & 44.4 \\
        \rowcolor{gray!10}
        FIS-D1 & \textbf{17.89} & 37.2 & 33.3 & 63.0 & 62.3 & 52.4 & 50.2 \\
        \rowcolor{gray!10}
        FIS-D2 & 14.89  & 37.6 & 33.2 & 63.3 & 63.8 & 53.7 & 51.6 \\
        \rowcolor{gray!10}
        FIS-D3 & 12.84 & \textbf{38.7} & \textbf{35.6} & \textbf{63.8} & \textbf{65.9} & \textbf{53.8} & \textbf{53.0} \\
        \hline
    \end{tabular}
    \caption{\textbf{Instance shadow detection on SOBA-testing.}
    Our FIS demonstrates a favorable trade-off between scaling and performance: while larger variants exhibit slower inference speeds, they achieve progressively higher accuracy.
    FIS-D1 surpasses SOTA methods across most evaluation metrics while achieving the fastest inference speed.
    Furthermore, FIS-D3 outperforms all existing approaches across all metrics, establishing new enhanced SOTA performance.
    Inference speed was measured using the entire SOBA-testing.}
    \label{tab:ap-soba-testing}
\end{table*}

\begin{table*}[t]
    \centering
    \begin{tabular}{l|cc|cc|cc}
        \hline
        Network & $SOAP_{segm}$ & $SOAP_{bbox}$ & Assoc. $AP_{segm}$ & Assoc. $AP_{bbox}$ & Inst. $AP_{segm}$ & Inst. $AP_{bbox}$ \\
        \hline
        LISA~\cite{bib:lisa}   & 10.4 & 10.1 & 20.7 & 25.8 & 23.8 & 24.3 \\
        SSIS~\cite{bib:ssis}   & 12.7 & 12.8 & 28.4 & 32.6 & 25.6 & 26.2 \\
        SSISv2~\cite{bib:ssisv2} & 17.7 & 15.1 & 34.6 & 37.3 & 31.0 & 28.4 \\
        \rowcolor{gray!10}
        FIS-D1 & 18.9 & 14.1 & 36.1 & 34.6 & 33.5 & 30.0 \\
        \rowcolor{gray!10}
        FIS-D2 & 20.6 & 16.7 & 38.5 & 35.9 & 35.6 & 32.7 \\
        \rowcolor{gray!10}
        FIS-D3 & \textbf{21.0} & \textbf{17.5} & \textbf{40.0} & \textbf{38.8} & \textbf{36.3} & \textbf{33.9} \\
        \hline
    \end{tabular}
    \caption{\textbf{Instance shadow detection on SOBA-challenge.}
    SOBA-challenge~\cite{bib:ssisv2} is a dataset for evaluating detection performance in complex scenarios.
    FIS-D3 establishes new enhanced SOTA performance even on this test set.
    }
    \label{tab:ap-soba-challenge}
\end{table*}

\subsubsection{Overall loss function}
The overall loss function for FIS can be expressed as:
\begin{equation}
    \mathcal{L} = \mathcal{L}_\text{IA-q} + \mathcal{L}_\text{pred} + \mathcal{L}'_\text{pred} + \mathcal{L}_\text{d},
\end{equation}
where $\mathcal{L}_\text{IA-q}$, $\mathcal{L}_\text{pred}$ and $\mathcal{L}'_\text{pred}$ are
instance activation loss, prediction loss and GT mask-guided loss, respectively.
$\mathcal{L}_\text{d}$ is shadow direction loss shown as \cref{equ:shadow-direction-loss}.
The Hungarian algorithm~\cite{bib:hungarian} is employed to
establish an optimal bipartite matching between predictions and the ground truth set similar to \cite{bib:detr, bib:hungarian-loss, bib:fastinst}.
Since detecting objects is more accurate than shadows,
this algorithm is performed considering only object masks, using the same costs as \cite{bib:fastinst}.\\
\textbf{Instance activation loss} $\mathcal{L}_\text{IA-q}$ is defined as:
\begin{equation}
    \mathcal{L}_\text{IA-q} = \lambda_\text{cls-q}\mathcal{L}_\text{cls-q},
\end{equation}
where $\lambda_\text{cls-q}$ is the loss weight and $\mathcal{L}_\text{cls-q}$ denotes the cross-entropy loss.
$\mathcal{L}_\text{cls-q}$ is calculated for the probabilities
obtained from the auxiliary classification head added on top of the feature map $E_4$.
Since each query in our model contains paired shadow-object features,
$\mathcal{L}_\text{cls-q}$ is calculated based on whether a query is assigned to a shadow-object pair.\\
\textbf{Prediction loss.} The $\mathcal{L}_\text{pred}$ is defined as:
\begin{equation}
    \begin{aligned}
        \mathcal{L}_{\text{pred}} = &\sum_{i=0}^{D} (\lambda_{\text{ce}} \mathcal{L}_{\text{o ce}}^{i} + \lambda_{\text{dice}} \mathcal{L}_{\text{o dice}}^{i} \\
        &+ \lambda_{\text{ce}} \mathcal{L}_{\text{s ce}}^{i} + \lambda_{\text{dice}} \mathcal{L}_{\text{s dice}}^{i}) + \lambda_{\text{cls}} \mathcal{L}_{\text{cls}}^{i}
    \end{aligned}
    \label{equ:prediction-loss}
\end{equation}
We extend the prediction loss from~\cite{bib:mask2former, bib:fastinst} to calculate loss for shadow masks,
where $D$ denotes the number of transformer decoder layers, $i = 0$ represents the prediction loss for IA-guided queries.
Here, $\mathcal{L}_{\text{o ce}}$ and $\mathcal{L}_{\text{o dice}}$ denote
the box-aware mask loss defined in \cref{equ:suppress-false-positive-loss} and the Dice loss~\cite{bib:dice-loss} for object masks, respectively.
Similarly, $\mathcal{L}_{\text{s ce}}$ and $\mathcal{L}_{\text{s dice}}$ denote the box-aware mask loss and Dice loss for shadow masks, respectively.
$\lambda_{\text{ce}}$, $\lambda_{\text{dice}}$ and $\lambda_{\text{cls}}$ are loss weights.
$\mathcal{L}_\text{cls}$ is calculated based on whether a query is assigned to a shadow-object pair,
similar to $\mathcal{L}_\text{cls-q}$.\\
\textbf{GT mask-guided loss.} $\mathcal{L}'_\text{pred}$ is similar to \cref{equ:prediction-loss}.
The only differences are that it applies mask-guided learning~\cite{bib:fastinst} while skipping calculation of the zeroth layer loss.

\begin{table*}[t]
    \centering
    \setlength{\tabcolsep}{4.5pt}
    \begin{tabular}{l|cc|cccccc}
    \hline
    & \multirow{3}{*}{\begin{tabular}{c} + shadow \\ direction \\ learning \end{tabular}}
    & \multirow{3}{*}{\begin{tabular}{c} + box \\ -aware \\ mask loss \end{tabular}}
    & \multirow{3}{*}{$SOAP_{segm}$}
    & \multirow{3}{*}{$SOAP_{bbox}$}
    & \multirow{3}{*}{\begin{tabular}{c} Assoc. \\ $AP_{segm}$ \end{tabular}}
    & \multirow{3}{*}{\begin{tabular}{c} Assoc. \\ $AP_{nbox}$ \end{tabular}}
    & \multirow{3}{*}{\begin{tabular}{c} Inst. \\ $AP_{segm}$ \end{tabular}}
    & \multirow{3}{*}{\begin{tabular}{c} Inst. \\ $AP_{bbox}$ \end{tabular}} \\
    &&&&&&&&\\
    &&&&&&&&\\
    \hline
    &  &  & 37.8 & 32.5 & 63.1 & 62.1 & 53.1 & 49.4 \\
    & $\checkmark$ &  & 38.1 & 32.8 & 63.5 & 62.2 & \textbf{53.8} & 50.6 \\
    &  & $\checkmark$ & 37.2 & 33.3 & 62.9 & 62.4 & 53.4 & 51.2 \\
    FIS-D3 & $\checkmark$ & $\checkmark$ & \textbf{38.7} & \textbf{35.6} & \textbf{63.8} & \textbf{65.9} & \textbf{53.8} & \textbf{53.0} \\
    \hline
    \end{tabular}
    \caption{\textbf{Training strategy analysis on SOBA-testing.}
    The results demonstrate that integrating the shadow direction learning strategy improves $SOAP_{segm}$ and $SOAP_{bbox}$.
    In addition, the box-aware mask loss particularly improves $SOAP_{bbox}$.
    By introducing both shadow direction learning and box-aware mask loss, we achieve the highest accuracy in all metrics.}
    \label{tab:ap-soba-testing-strategy}
\end{table*}

\begin{table*}[h]
    \centering
    \setlength{\tabcolsep}{4.5pt}
    \begin{tabular}{l|cc|cccccc}
    \hline
    & \multirow{3}{*}{\begin{tabular}{c} + shadow \\ direction \\ learning \end{tabular}}
    & \multirow{3}{*}{\begin{tabular}{c} + box \\ -aware \\ mask loss \end{tabular}}
    & \multirow{3}{*}{$SOAP_{segm}$}
    & \multirow{3}{*}{$SOAP_{bbox}$}
    & \multirow{3}{*}{\begin{tabular}{c} Assoc. \\ $AP_{segm}$ \end{tabular}}
    & \multirow{3}{*}{\begin{tabular}{c} Assoc. \\ $AP_{bbox}$ \end{tabular}}
    & \multirow{3}{*}{\begin{tabular}{c} Inst. \\ $AP_{segm}$ \end{tabular}}
    & \multirow{3}{*}{\begin{tabular}{c} Inst. \\ $AP_{bbox}$ \end{tabular}} \\
    &&&&&&&&\\
    &&&&&&&&\\
    \hline
    &  &  & 18.4 & 14.0 & 36.1 & 32.1 & 33.6 & 29.1 \\
    & $\checkmark$ &  & 20.5 & 15.6 & 38.0 & 34.8 & 35.3 & 31.2 \\
    &  & $\checkmark$ & 19.7 & 16.2 & 38.6 & 35.6 & 35.4 & 32.9 \\
    FIS-D3 & $\checkmark$ & $\checkmark$ & \textbf{21.0} & \textbf{17.5} & \textbf{40.0} & \textbf{38.8} & \textbf{36.3} & \textbf{33.9} \\
    \hline
    \end{tabular}
    \caption{\textbf{Training strategy analysis on SOBA-challenge.} The effectiveness of the two strategies is also demonstrated even on this test set.}
    \label{tab:ap-soba-challenge-strategy}
\end{table*}

\section{Experiments}
This section compares the performance of FIS using the 
SOBA~\cite{bib:lisa, bib:ssis, bib:ssisv2} dataset with
other SOTA instance shadow detection methods.
In the evaluation, relations between the inference speed and the detection
accuracy are shown while changing the number of association transformer decoder layers implemented in FIS.
FIS-D$\Gamma$ represents FIS with $\Gamma$ association transformer decoder layers.
In addition to these evaluations,
we also demonstrate the effectiveness of the proposed training strategy
composed of shadow direction learning and box-aware mask loss.

\subsection{Implementation details}
We used the AdamW~\cite{bib:adamw} optimizer with a learning rate of $1 \times 10^{-4}$ and a weight decay of $0.05$.
There is a $1,000$ iterations linear warmup and a cosine decaying schedule afterward.
Our model was trained for $100$ epochs with a batch size of $4$.
We employed the same data augmentation method as SSISv2~\cite{bib:ssisv2}.
We initialized FIS weights by transferring the applicable portions from FastInst pretrained on the COCO~\cite{bib:coco} dataset.
To avoid data leakage, we excluded images from the COCO dataset that overlapped with the SOBA-testing for pretraining.
The loss weights $\lambda_{\text{cls-q}}$, $\lambda_{\text{cls}}$, $\lambda_{\text{ce}}$, $\lambda_{\text{dice}}$ and $\lambda_{d}$
were set to 20.0, 2.0, 5.0, 5.0 and 0.1, respectively.
For shadow direction learning, $\beta$ was set to $0.5$.
For box-aware mask loss, $N$ and $\alpha$ were set to $75$ and $50$, respectively.
Following the evaluation protocol of existing methods~\cite{bib:lisa, bib:ssis, bib:ssisv2},
we reported SOAP~\cite{bib:lisa}, instance AP, and association AP on both SOBA-testing and SOBA-challenge for the model trained on SOBA-training.
Since FIS does not have a mechanism to detect bounding boxes,
the $AP_{bbox}$ evaluation of FIS used the AABB of the generated mask.
We also measured the inference speed of the shadow and object masks as fps.
All inference speed measurements were conducted on an RTX3090 GPU with a batch size of $1$.
Unless specified, we resized input images to a shorter edge of 800 pixels and ensured a maximum length of 1,333 pixels
during testing, similar to existing methods~\cite{bib:lisa, bib:ssis, bib:ssisv2}.

\subsection{Comparison with state-of-the-art methods}
We compared our proposed FIS with SOTA methods.
The comparative results for SOBA-testing and SOBA-challenge are presented in \cref{tab:ap-soba-testing} and \ref{tab:ap-soba-challenge}, respectively.
All FIS series surpassed SOTA methods in instance AP, which evaluates individual shadow and object detection accuracy.
This demonstrates that our simple architectural extension preserves the high detection capability of the base FastInst model.
Almost FIS series also surpassed SOTA methods in SOAP, which evaluates shadow-object detection accuracy and correct pairing performance.
Existing instance shadow detection methods employ additional detection and heuristic pairing algorithms for pairing,
whereas our method does not use pairing process, detecting shadows and objects as pairs.
These SOAP results demonstrate the effectiveness of our paired detection method.
Almost FIS series surpassed SOTA methods in association AP, which evaluates detection of integrated regions for each shadow-object pair.
Remarkably, FIS-D3 exceeded SOTA performance across all metrics on SOBA-testing and SOBA-challenge.

As shown in the second column of \cref{tab:ap-soba-testing}, FIS series demonstrates high inference speed.
Remarkably, FIS-D1 achieves higher inference speed than all existing methods.
This performance advantage is a result of our simple extension to the FastInst, real-time instance segmentation model.

\begin{figure*}[t]
    \centering
    \begin{minipage}[c]{\linewidth}
        \begin{minipage}[c]{0.19\linewidth}
            \centering
            \includegraphics[width=\linewidth]{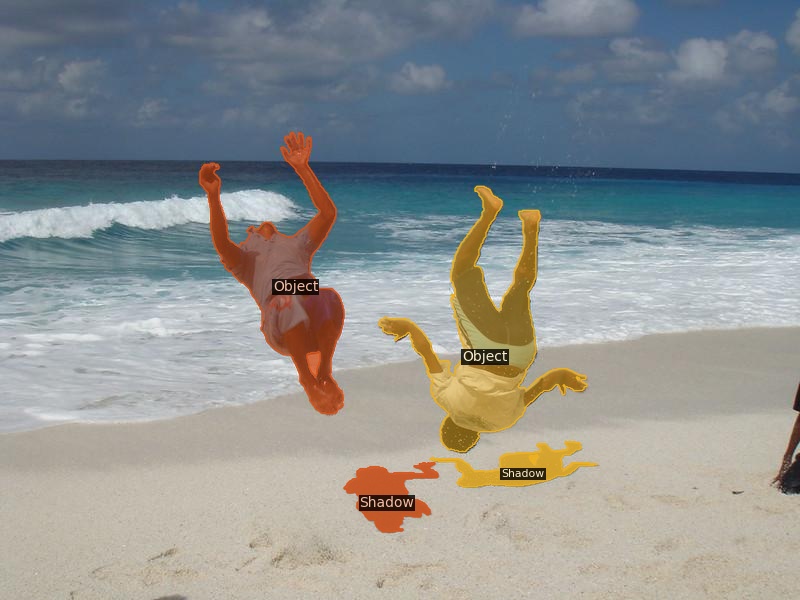}
        \end{minipage}
        \hfill
        \begin{minipage}[c]{0.19\linewidth}
            \centering
            \includegraphics[width=\linewidth]{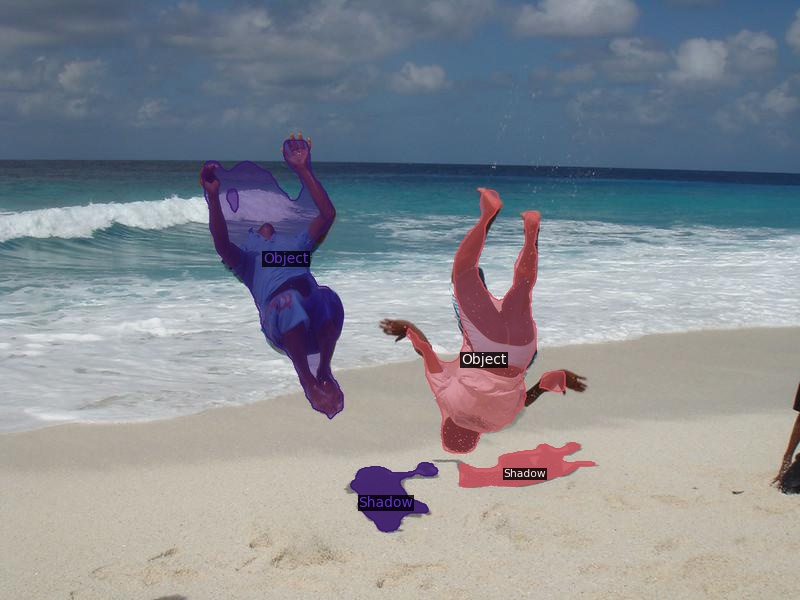}
        \end{minipage}
        \hfill
        \begin{minipage}[c]{0.19\linewidth}
            \centering
            \includegraphics[width=\linewidth]{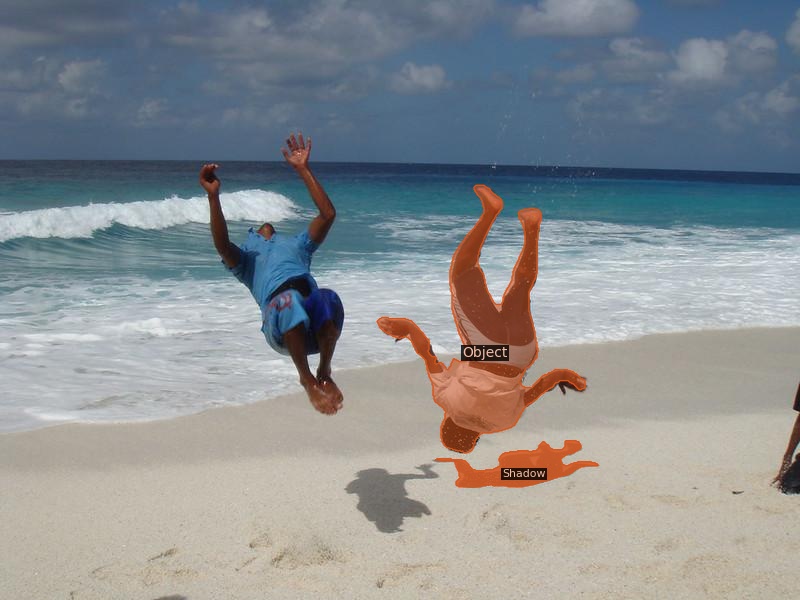}
        \end{minipage}
        \hfill
        \begin{minipage}[c]{0.19\linewidth}
            \centering
            \includegraphics[width=\linewidth]{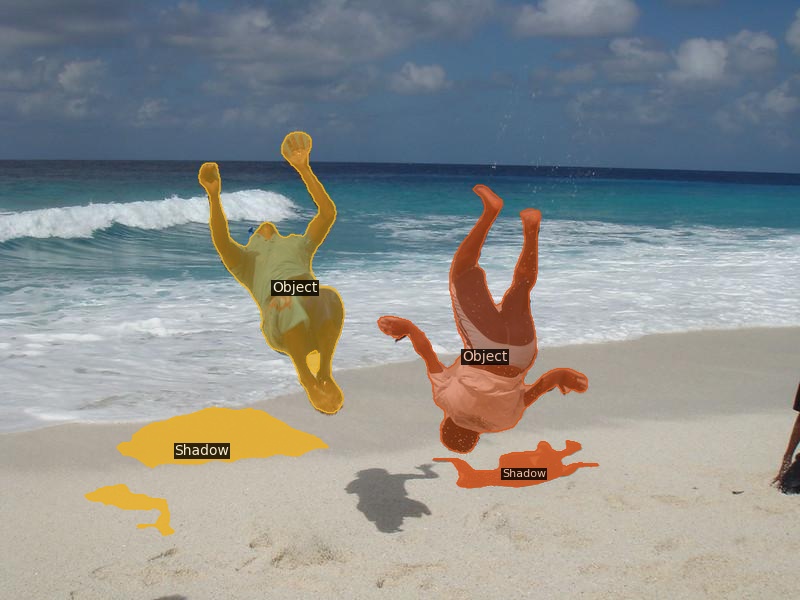}
        \end{minipage}
        \hfill
        \begin{minipage}[c]{0.19\linewidth}
            \centering
            \includegraphics[width=\linewidth]{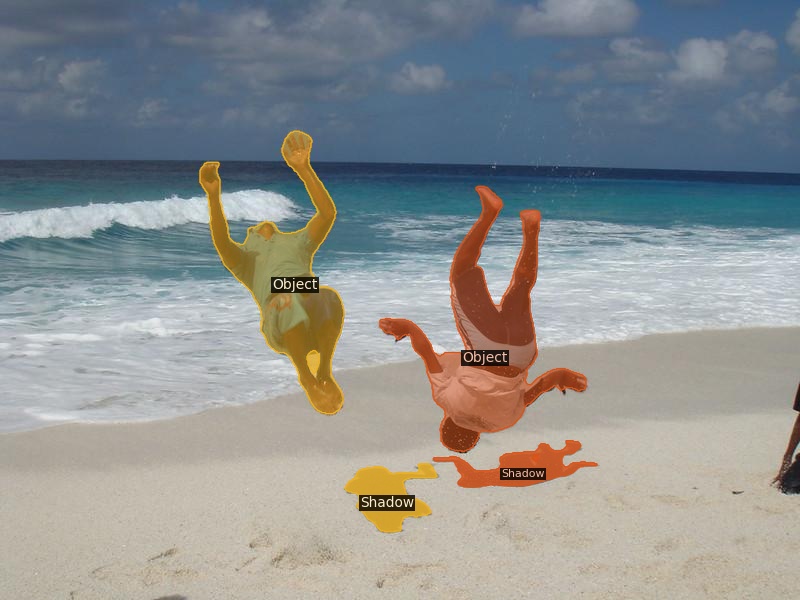}
        \end{minipage}
    \end{minipage}
    
    \vspace{0.8em}
    
    \begin{minipage}[c]{\linewidth}
        \begin{minipage}[c]{0.19\linewidth}
            \centering
            \includegraphics[width=\linewidth]{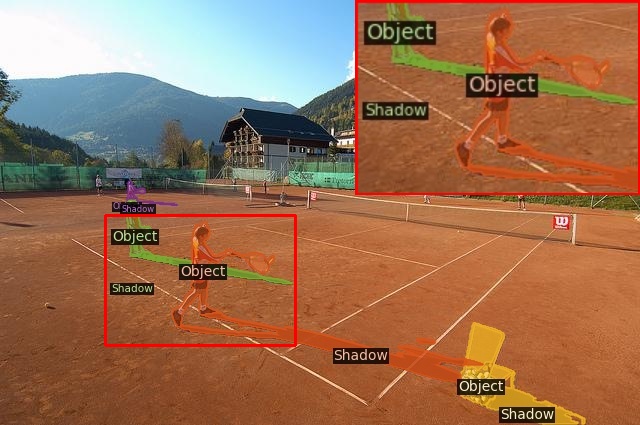}
        \end{minipage}
        \hfill
        \begin{minipage}[c]{0.19\linewidth}
            \centering
            \includegraphics[width=\linewidth]{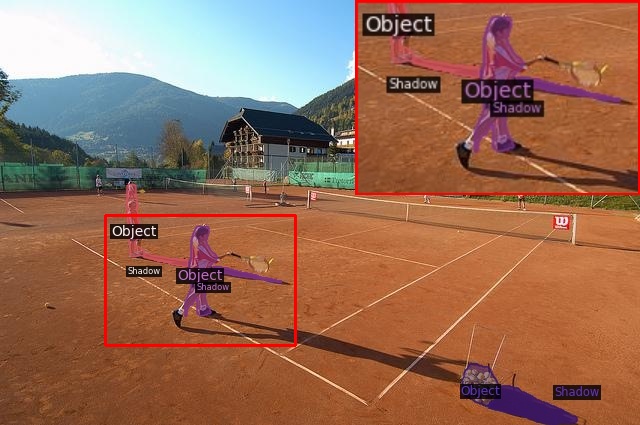}
        \end{minipage}
        \hfill
        \begin{minipage}[c]{0.19\linewidth}
            \centering
            \includegraphics[width=\linewidth]{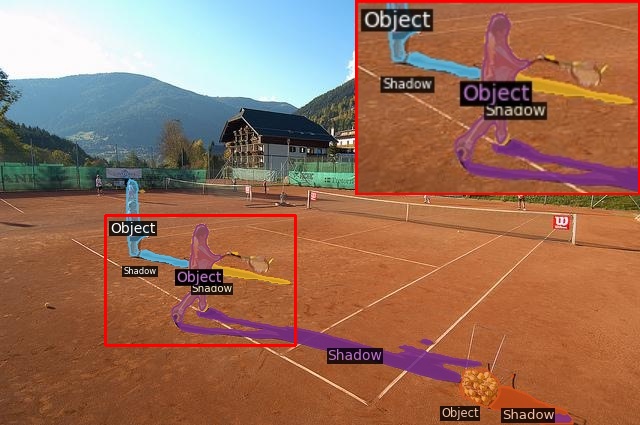}
        \end{minipage}
        \hfill
        \begin{minipage}[c]{0.19\linewidth}
            \centering
            \includegraphics[width=\linewidth]{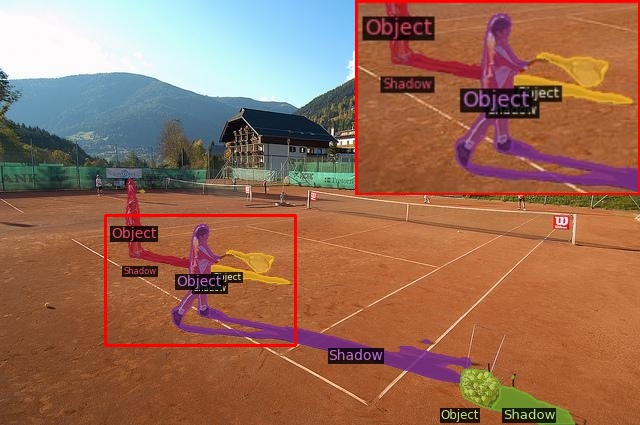}
        \end{minipage}
        \hfill
        \begin{minipage}[c]{0.19\linewidth}
            \centering
            \includegraphics[width=\linewidth]{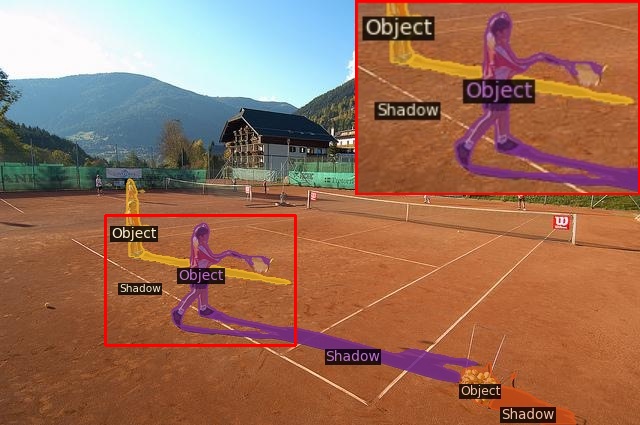}
        \end{minipage}
    \end{minipage}
    
    \vspace{0.8em}
    
    \begin{minipage}[c]{\linewidth}
        \begin{minipage}[c]{0.19\linewidth}
            \centering
            \includegraphics[width=\linewidth]{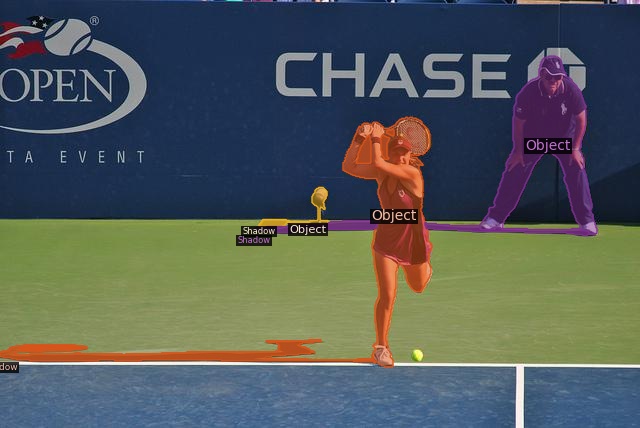}
        \end{minipage}
        \hfill
        \begin{minipage}[c]{0.19\linewidth}
            \centering
            \includegraphics[width=\linewidth]{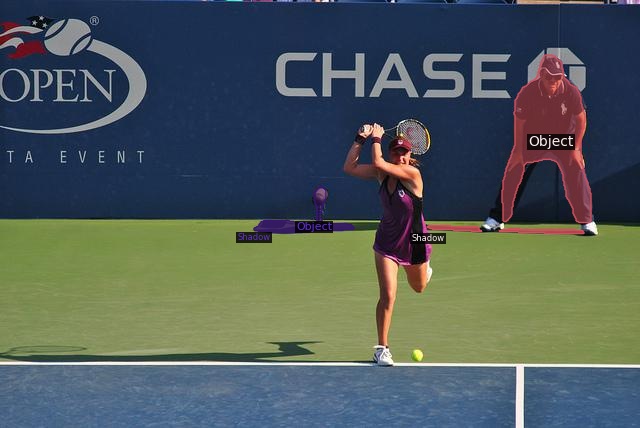}
        \end{minipage}
        \hfill
        \begin{minipage}[c]{0.19\linewidth}
            \centering
            \includegraphics[width=\linewidth]{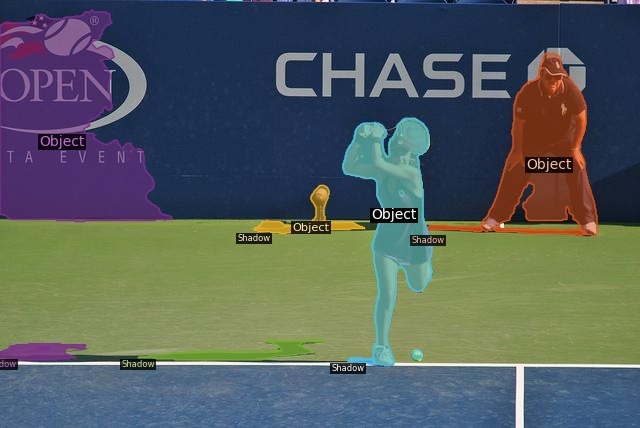}
        \end{minipage}
        \hfill
        \begin{minipage}[c]{0.19\linewidth}
            \centering
            \includegraphics[width=\linewidth]{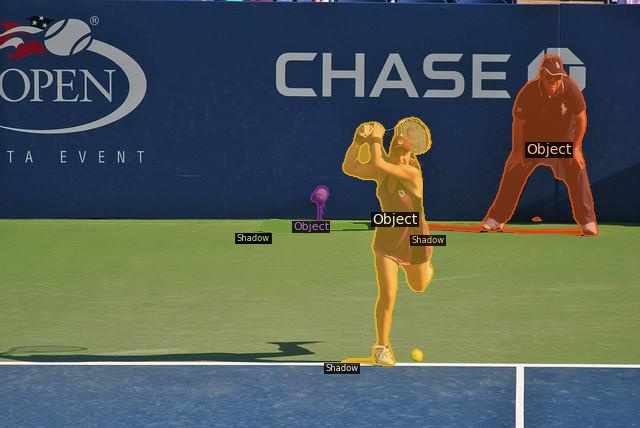}
        \end{minipage}
        \hfill
        \begin{minipage}[c]{0.19\linewidth}
            \centering
            \includegraphics[width=\linewidth]{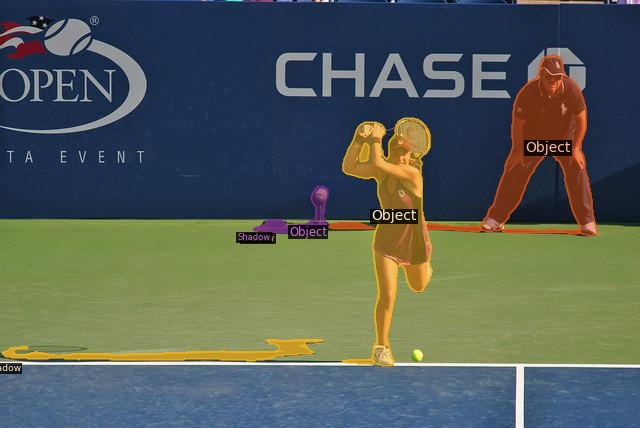}
        \end{minipage}
    \end{minipage}
    
    \vspace{0.8em}
    
    \begin{minipage}[c]{\linewidth}
        \begin{minipage}[c]{0.19\linewidth}
            \centering
            \includegraphics[width=\linewidth]{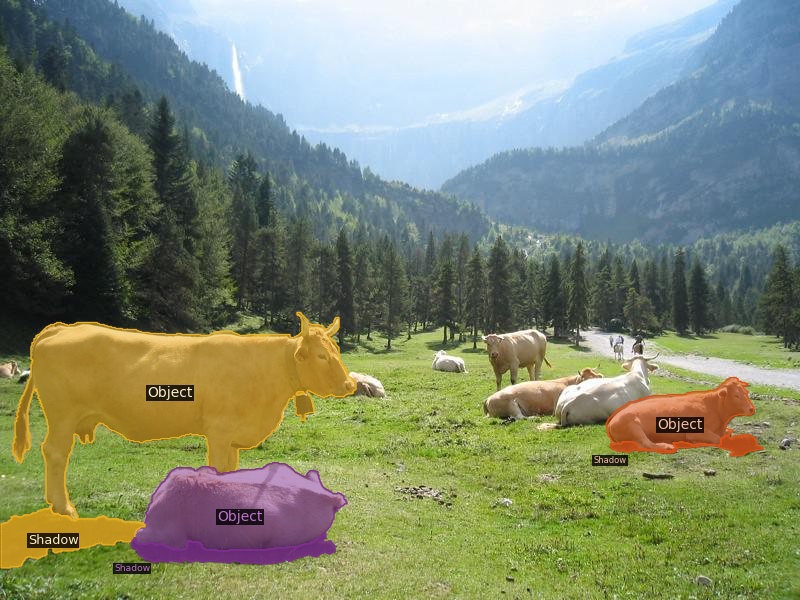}
        \end{minipage}
        \hfill
        \begin{minipage}[c]{0.19\linewidth}
            \centering
            \includegraphics[width=\linewidth]{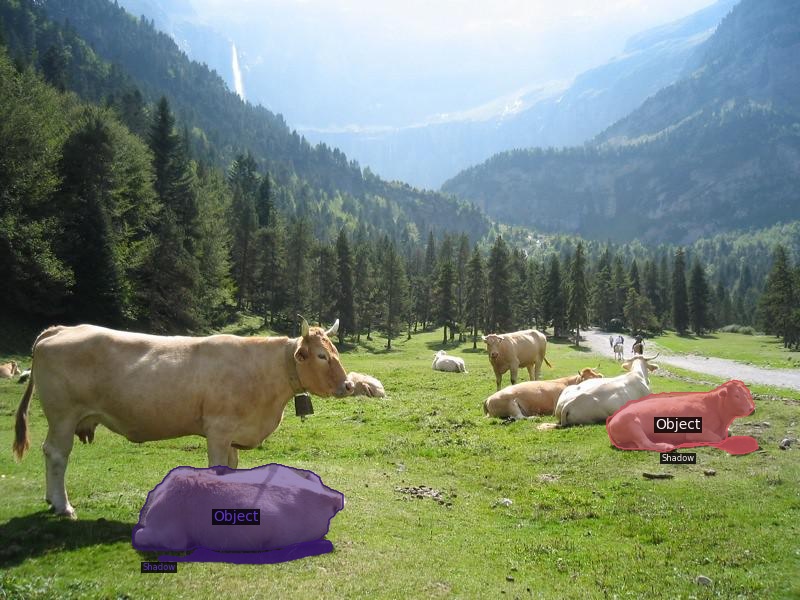}
        \end{minipage}
        \hfill
        \begin{minipage}[c]{0.19\linewidth}
            \centering
            \includegraphics[width=\linewidth]{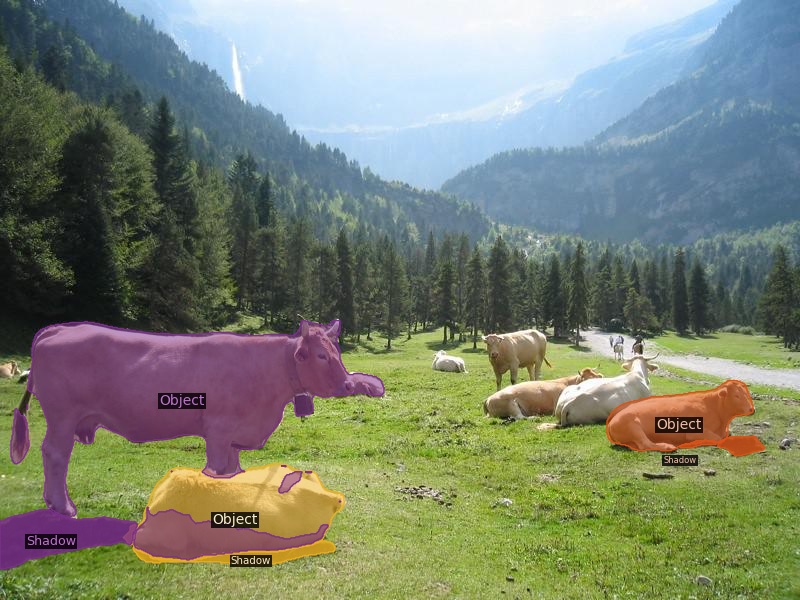}
        \end{minipage}
        \hfill
        \begin{minipage}[c]{0.19\linewidth}
            \centering
            \includegraphics[width=\linewidth]{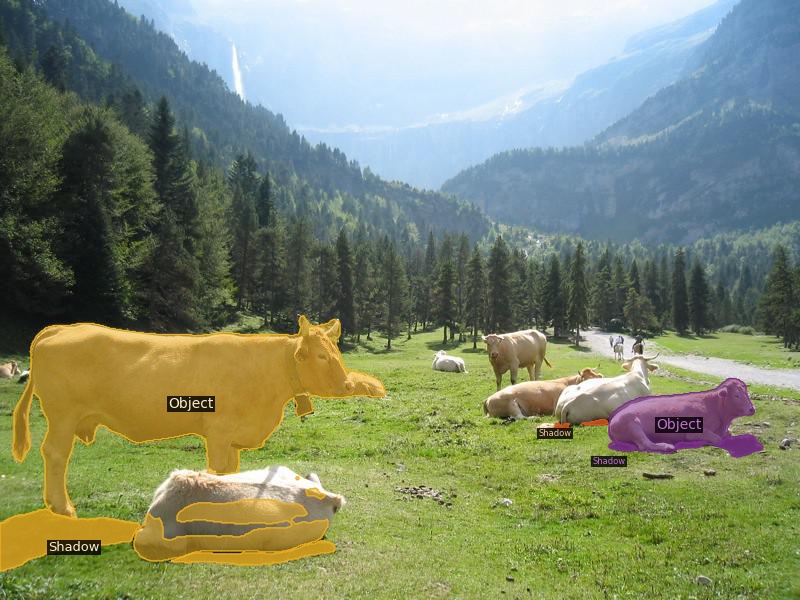}
        \end{minipage}
        \hfill
        \begin{minipage}[c]{0.19\linewidth}
            \centering
            \includegraphics[width=\linewidth]{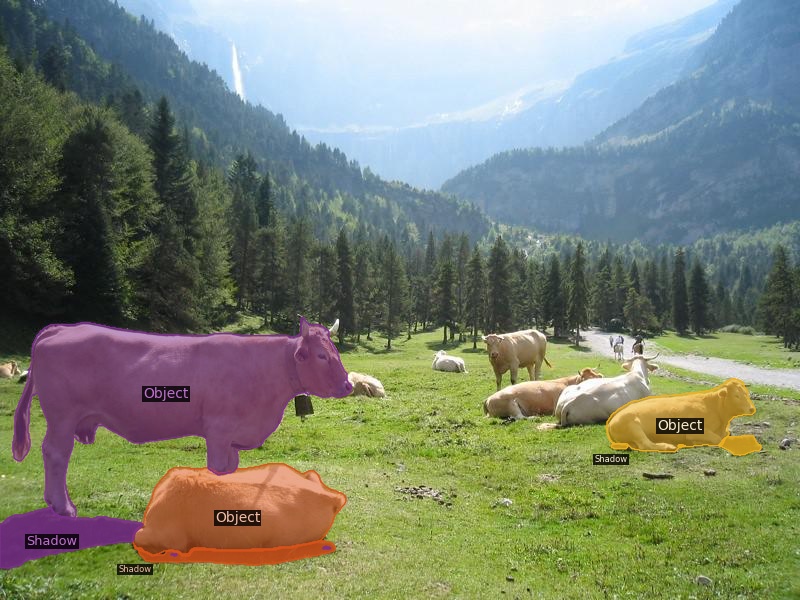}
        \end{minipage}
    \end{minipage}
    
    \vspace{0.8em}
    
    \begin{minipage}[c]{\linewidth}
        \begin{minipage}[c]{0.19\linewidth}
            \centering
            \includegraphics[width=\linewidth]{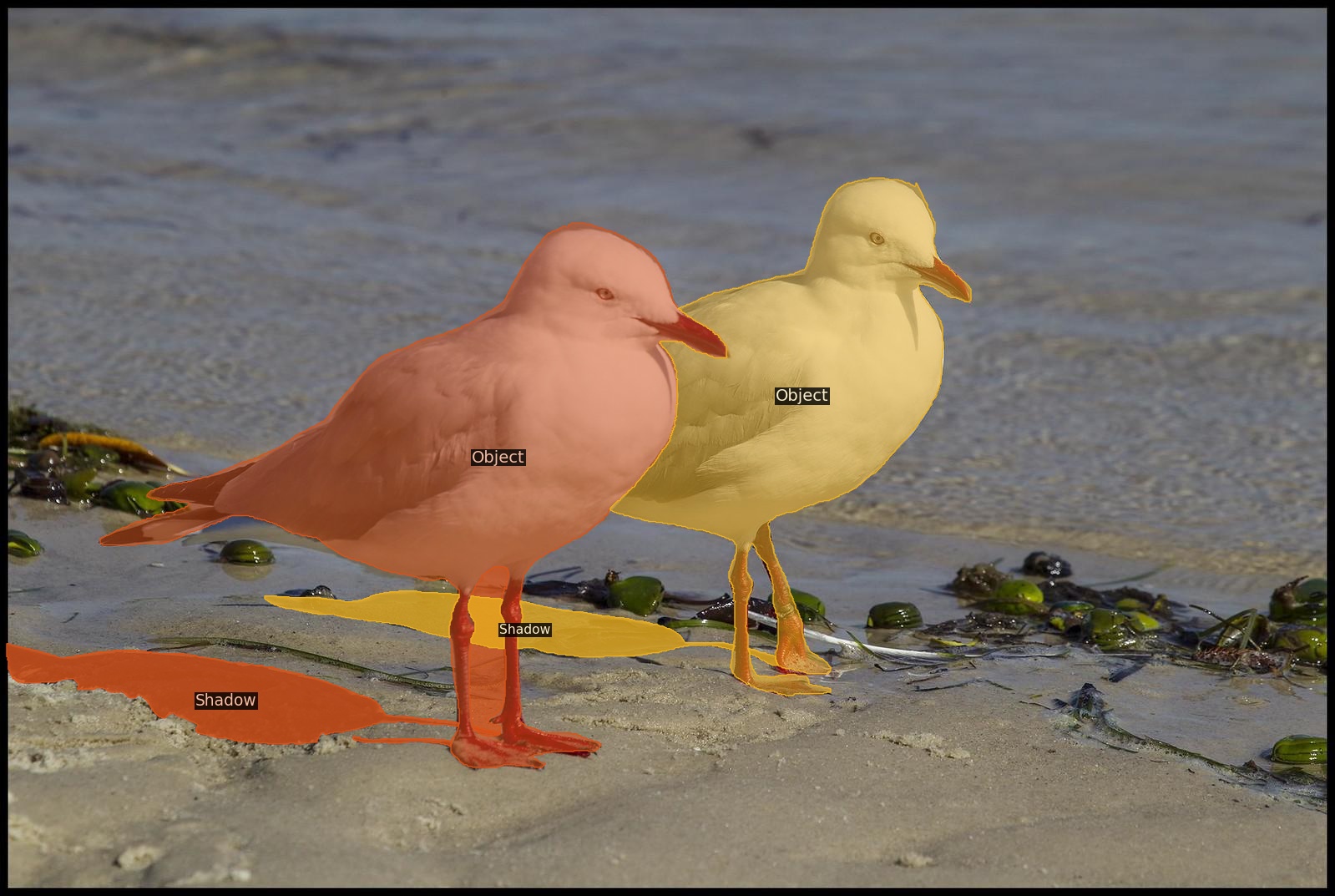}
        \end{minipage}
        \hfill
        \begin{minipage}[c]{0.19\linewidth}
            \centering
            \includegraphics[width=\linewidth]{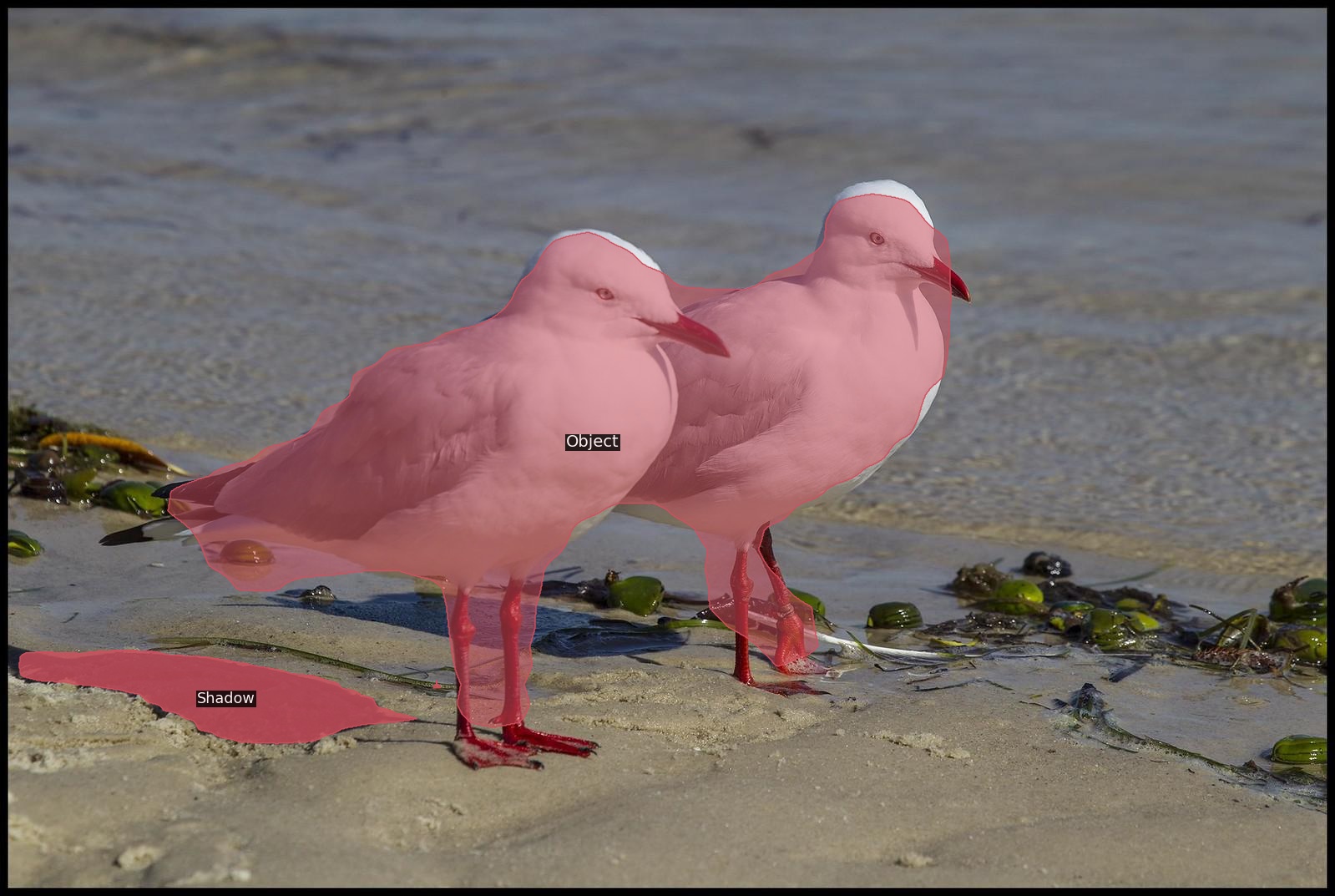}
        \end{minipage}
        \hfill
        \begin{minipage}[c]{0.19\linewidth}
            \centering
            \includegraphics[width=\linewidth]{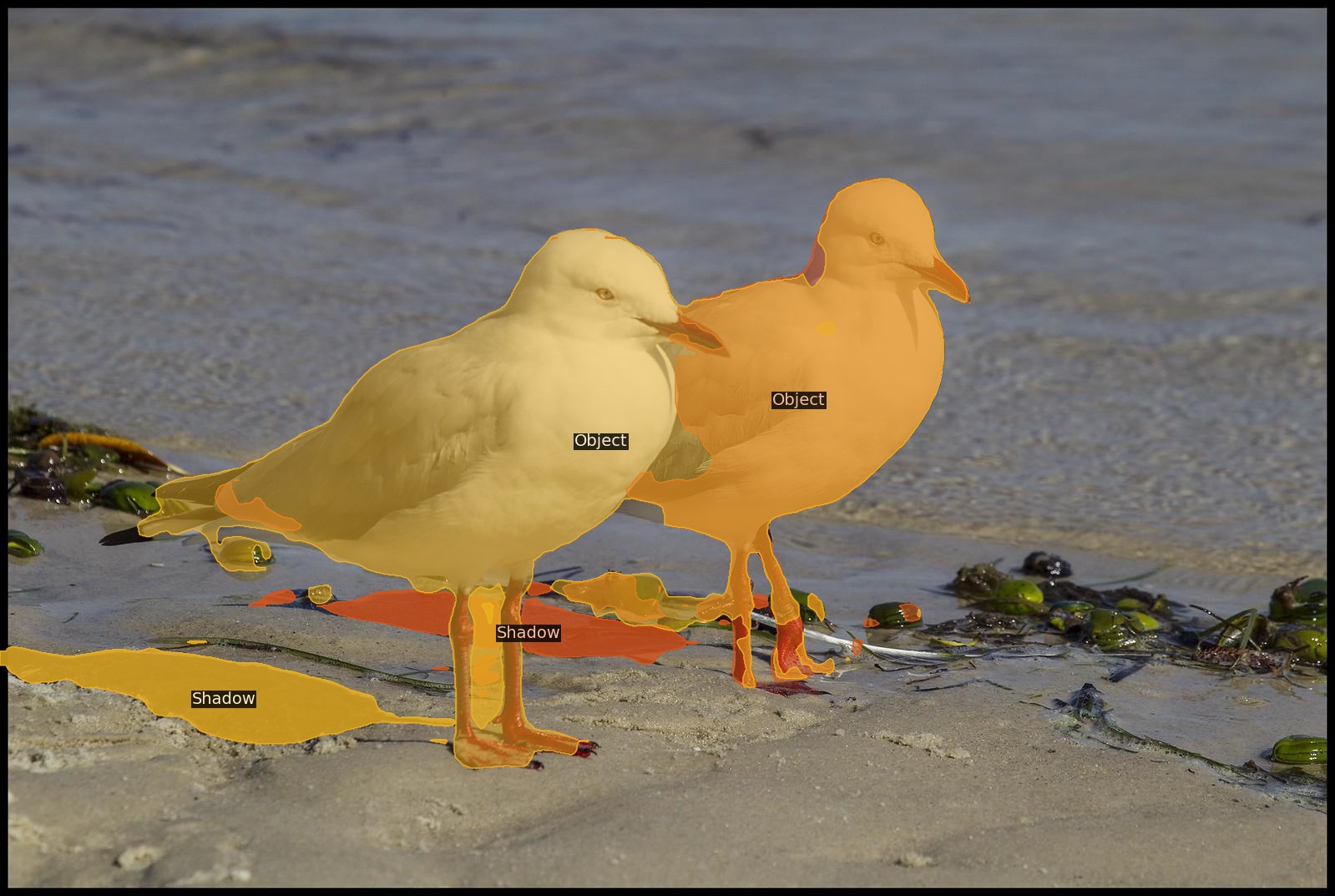}
        \end{minipage}
        \hfill
        \begin{minipage}[c]{0.19\linewidth}
            \centering
            \includegraphics[width=\linewidth]{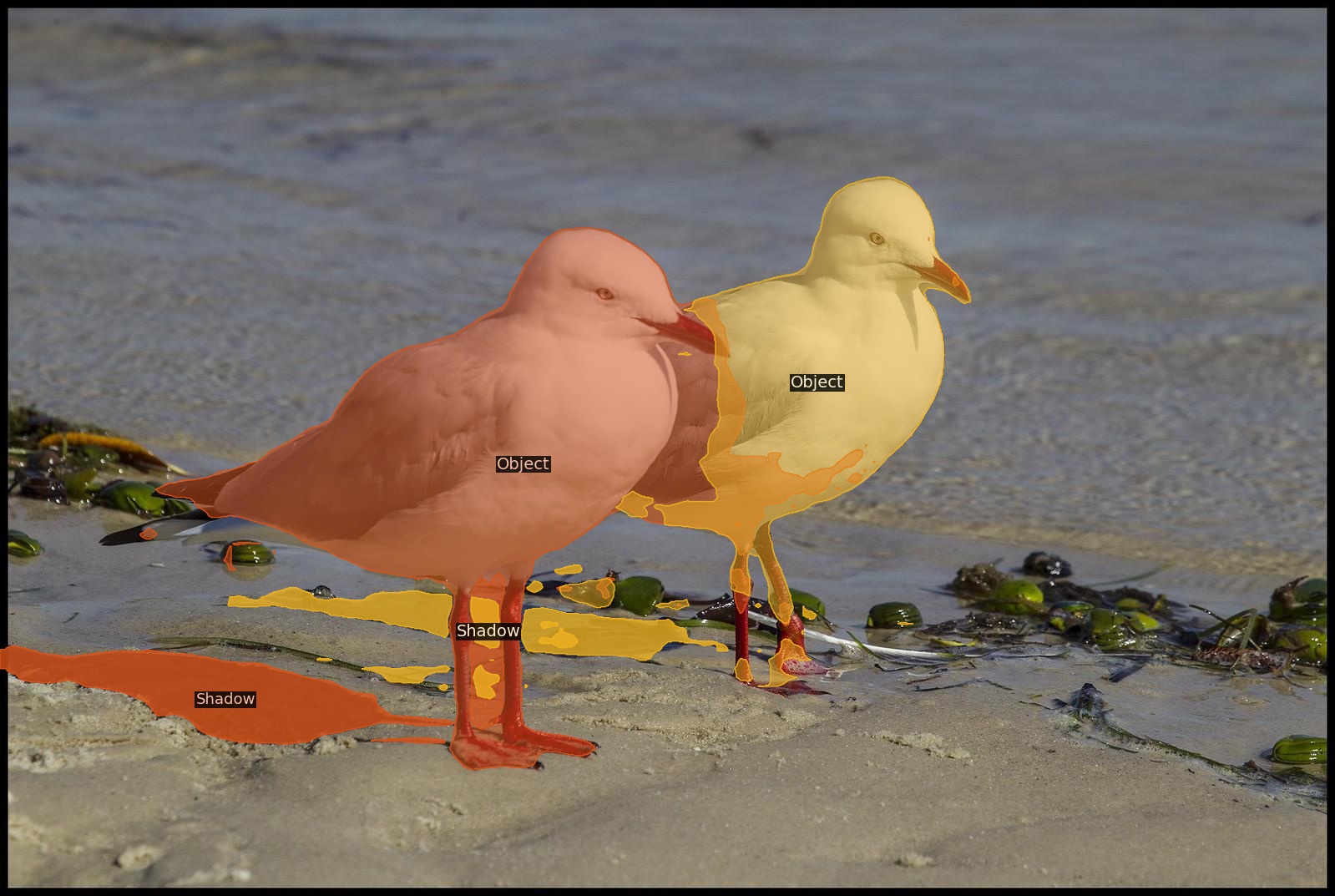}
        \end{minipage}
        \hfill
        \begin{minipage}[c]{0.19\linewidth}
            \centering
            \includegraphics[width=\linewidth]{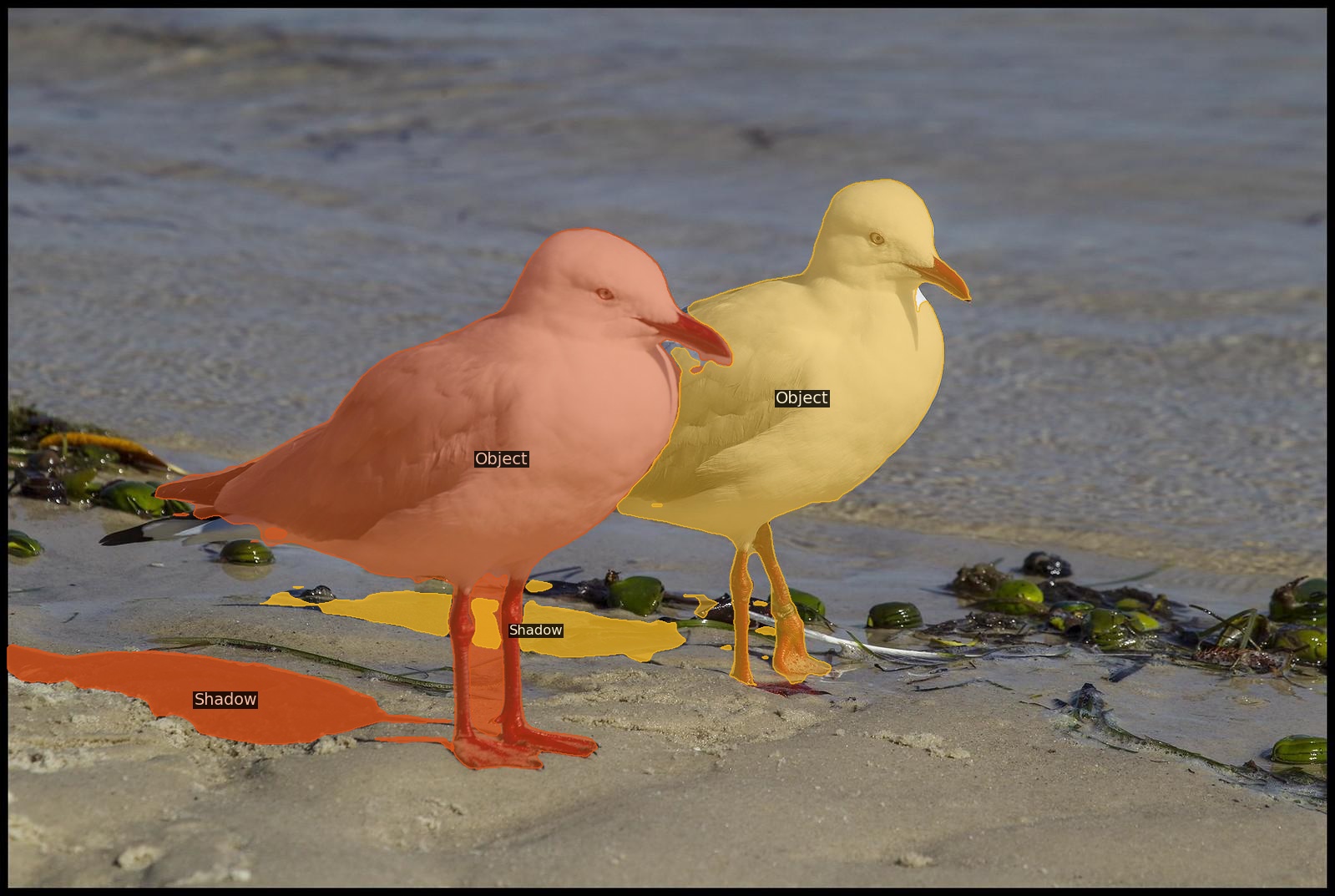}
        \end{minipage}
    \end{minipage}
    
    \vspace{0.8em}
    
    \begin{minipage}[c]{\linewidth}
        \begin{minipage}[c]{0.19\linewidth}
            \centering
            \includegraphics[width=\linewidth]{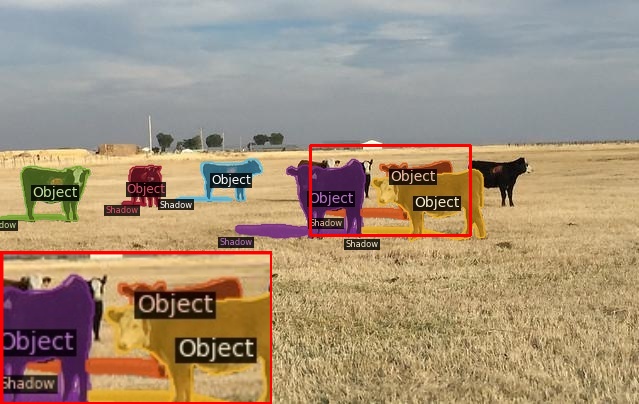}
        \end{minipage}
        \hfill
        \begin{minipage}[c]{0.19\linewidth}
            \centering
            \includegraphics[width=\linewidth]{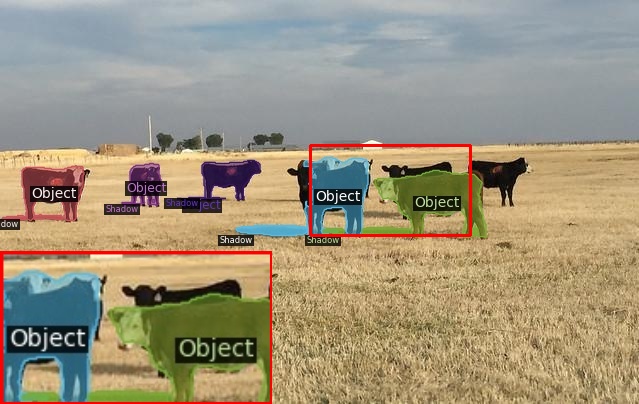}
        \end{minipage}
        \hfill
        \begin{minipage}[c]{0.19\linewidth}
            \centering
            \includegraphics[width=\linewidth]{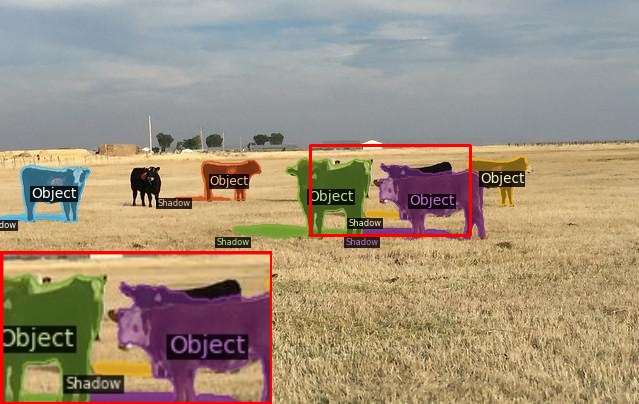}
        \end{minipage}
        \hfill
        \begin{minipage}[c]{0.19\linewidth}
            \centering
            \includegraphics[width=\linewidth]{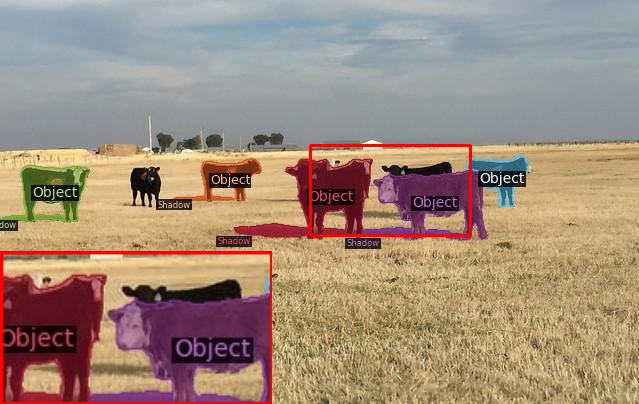}
        \end{minipage}
        \hfill
        \begin{minipage}[c]{0.19\linewidth}
            \centering
            \includegraphics[width=\linewidth]{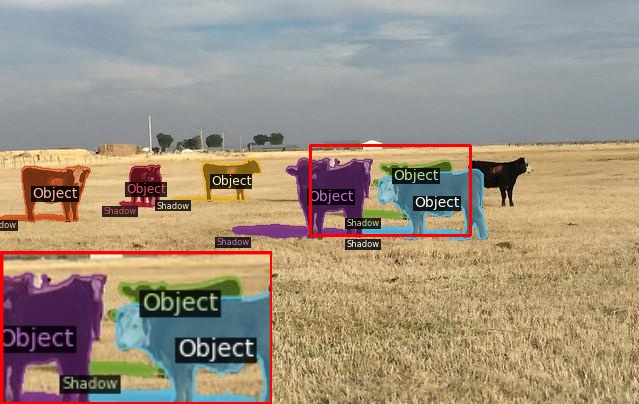}
        \end{minipage}
    \end{minipage}
    
    \vspace{0.8em}
    
    \begin{minipage}[c]{\linewidth}
        \begin{minipage}[c]{0.19\linewidth}
            \centering
            \includegraphics[width=\linewidth]{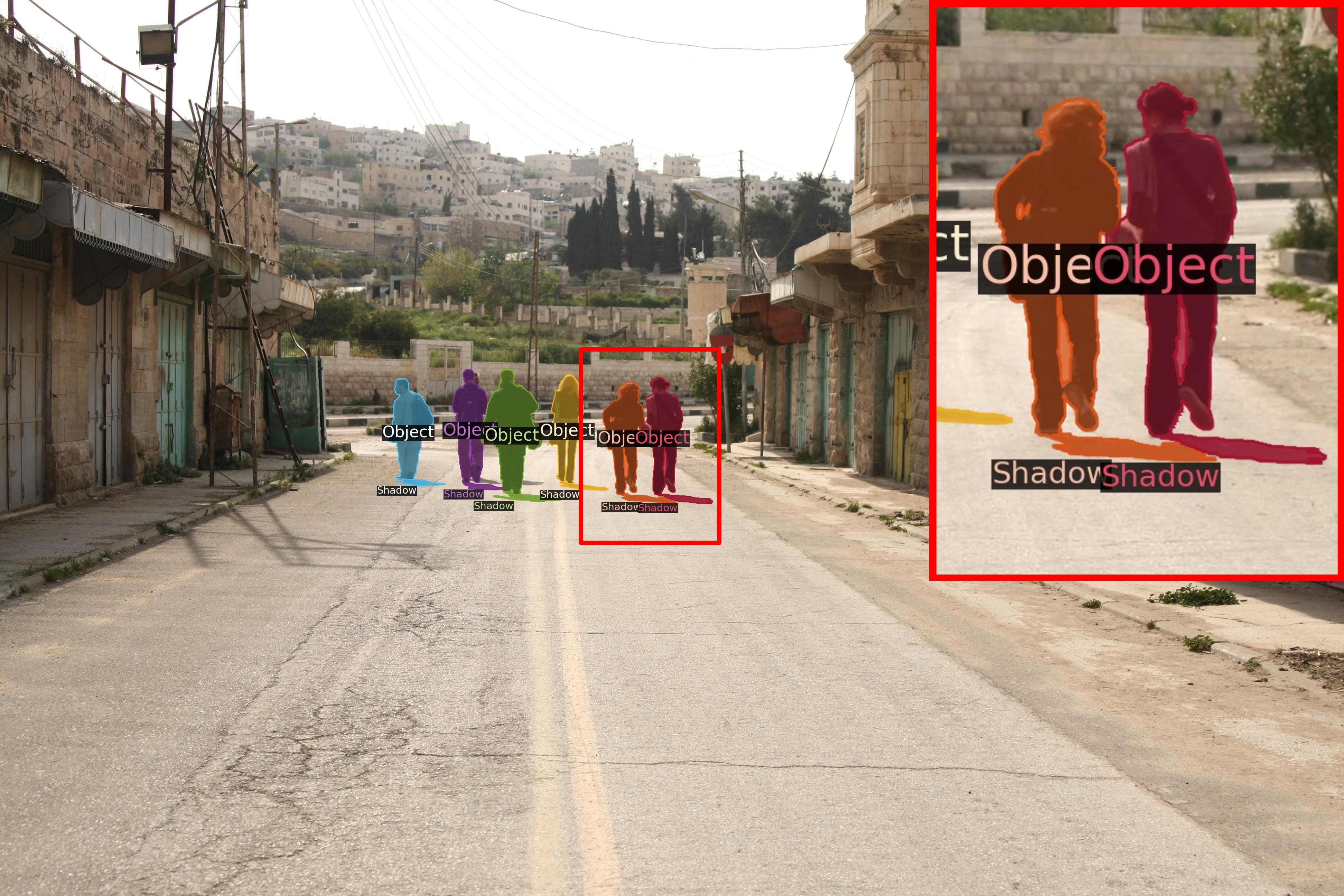}
        \end{minipage}
        \hfill
        \begin{minipage}[c]{0.19\linewidth}
            \centering
            \includegraphics[width=\linewidth]{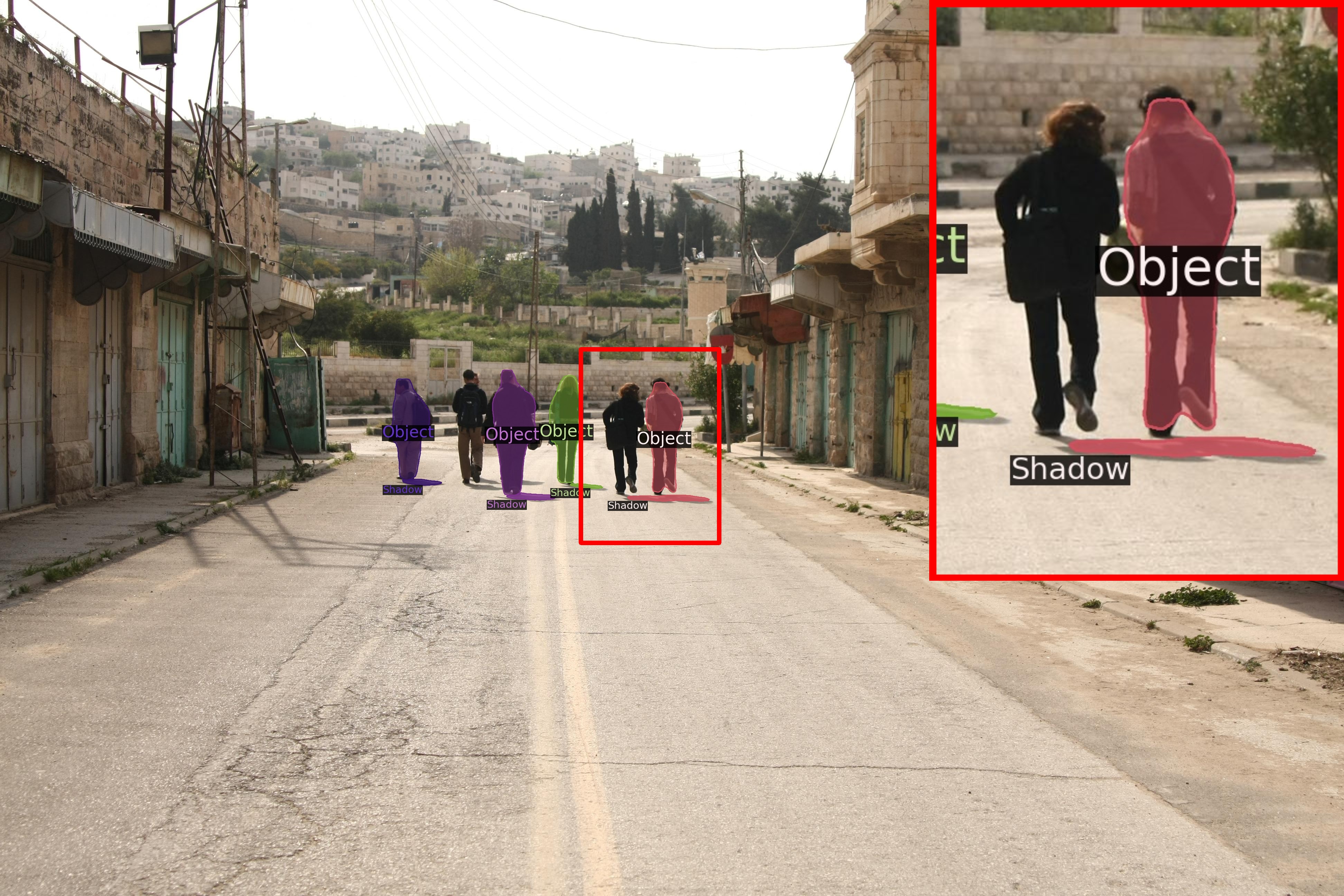}
        \end{minipage}
        \hfill
        \begin{minipage}[c]{0.19\linewidth}
            \centering
            \includegraphics[width=\linewidth]{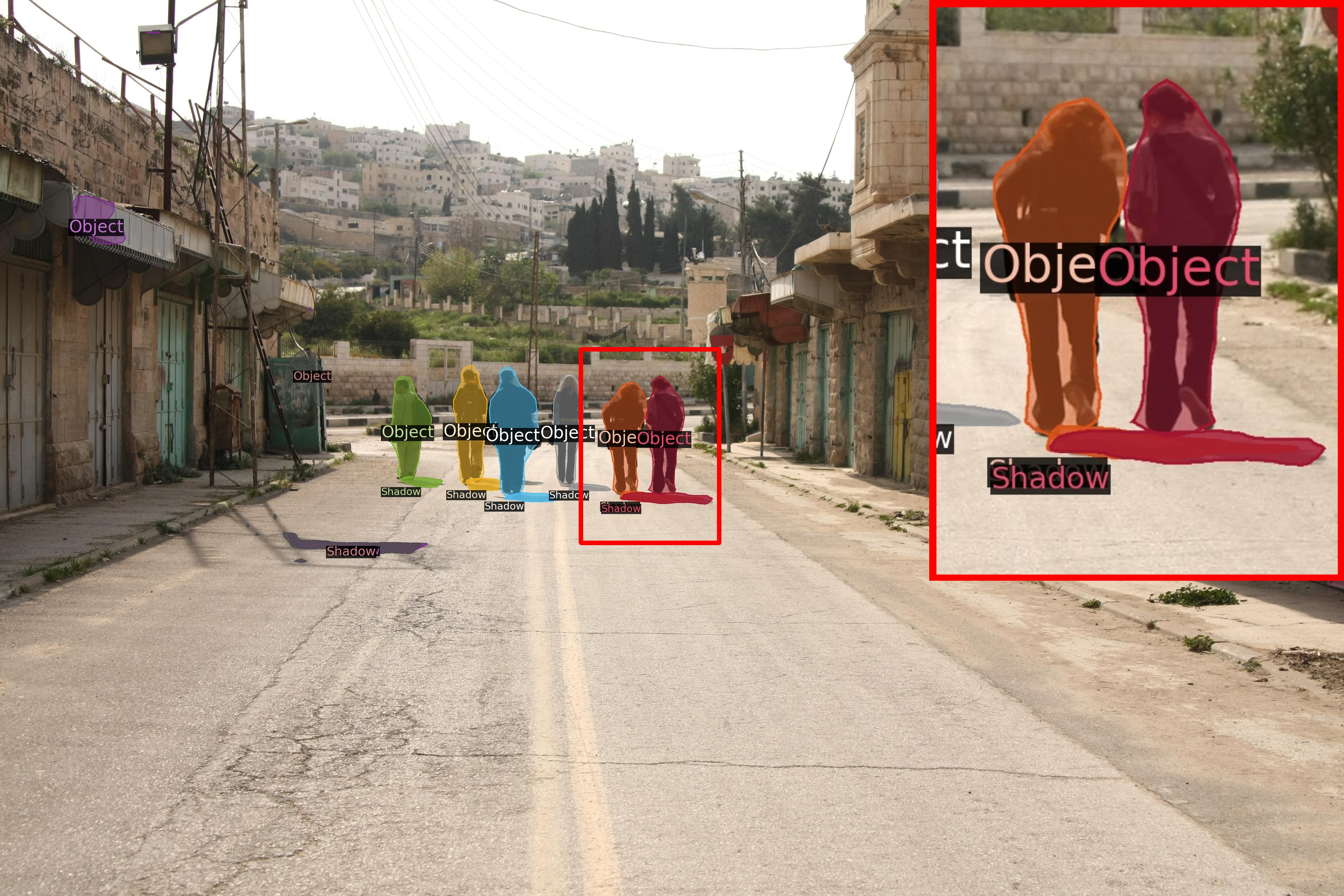}
        \end{minipage}
        \hfill
        \begin{minipage}[c]{0.19\linewidth}
            \centering
            \includegraphics[width=\linewidth]{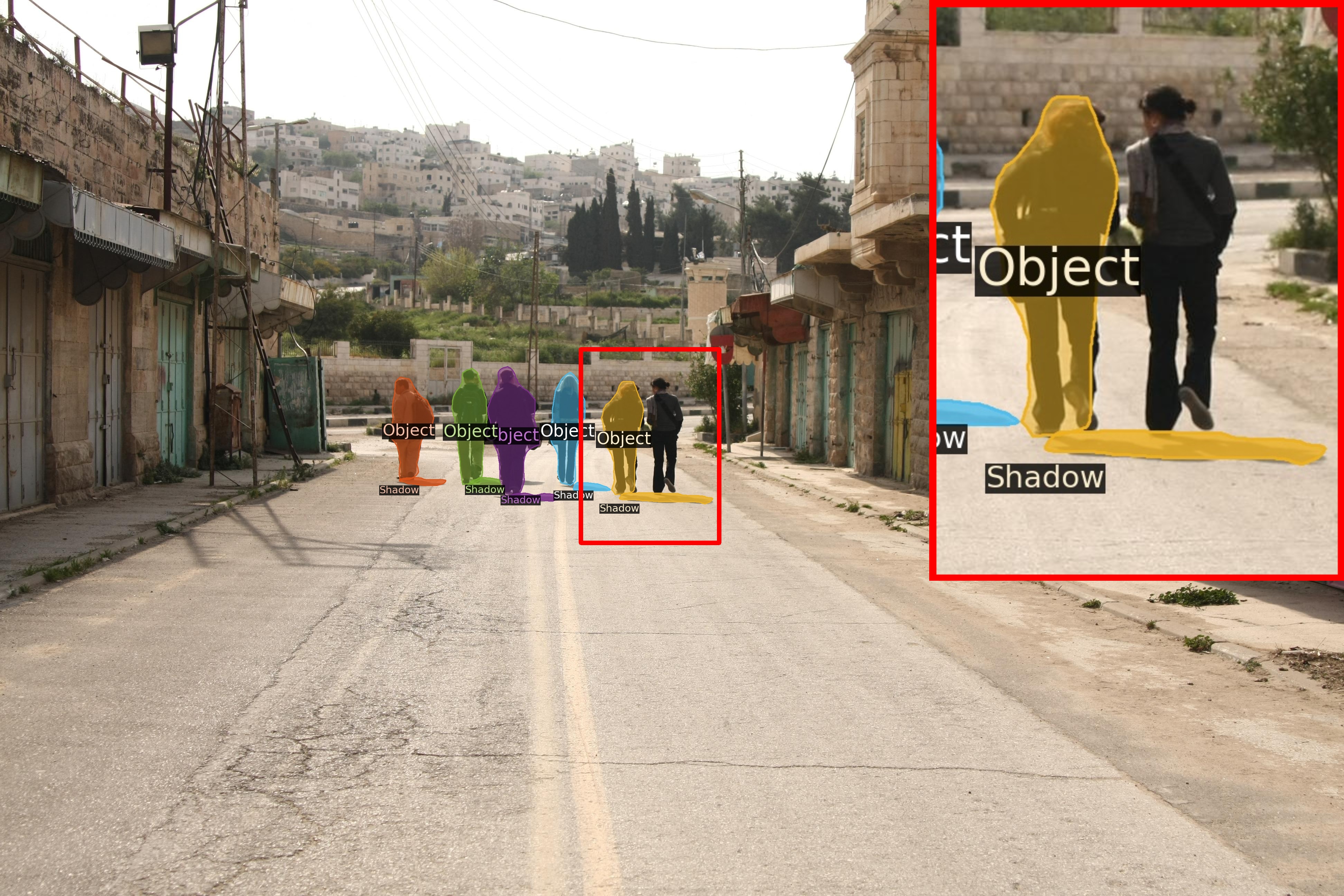}
        \end{minipage}
        \hfill
        \begin{minipage}[c]{0.19\linewidth}
            \centering
            \includegraphics[width=\linewidth]{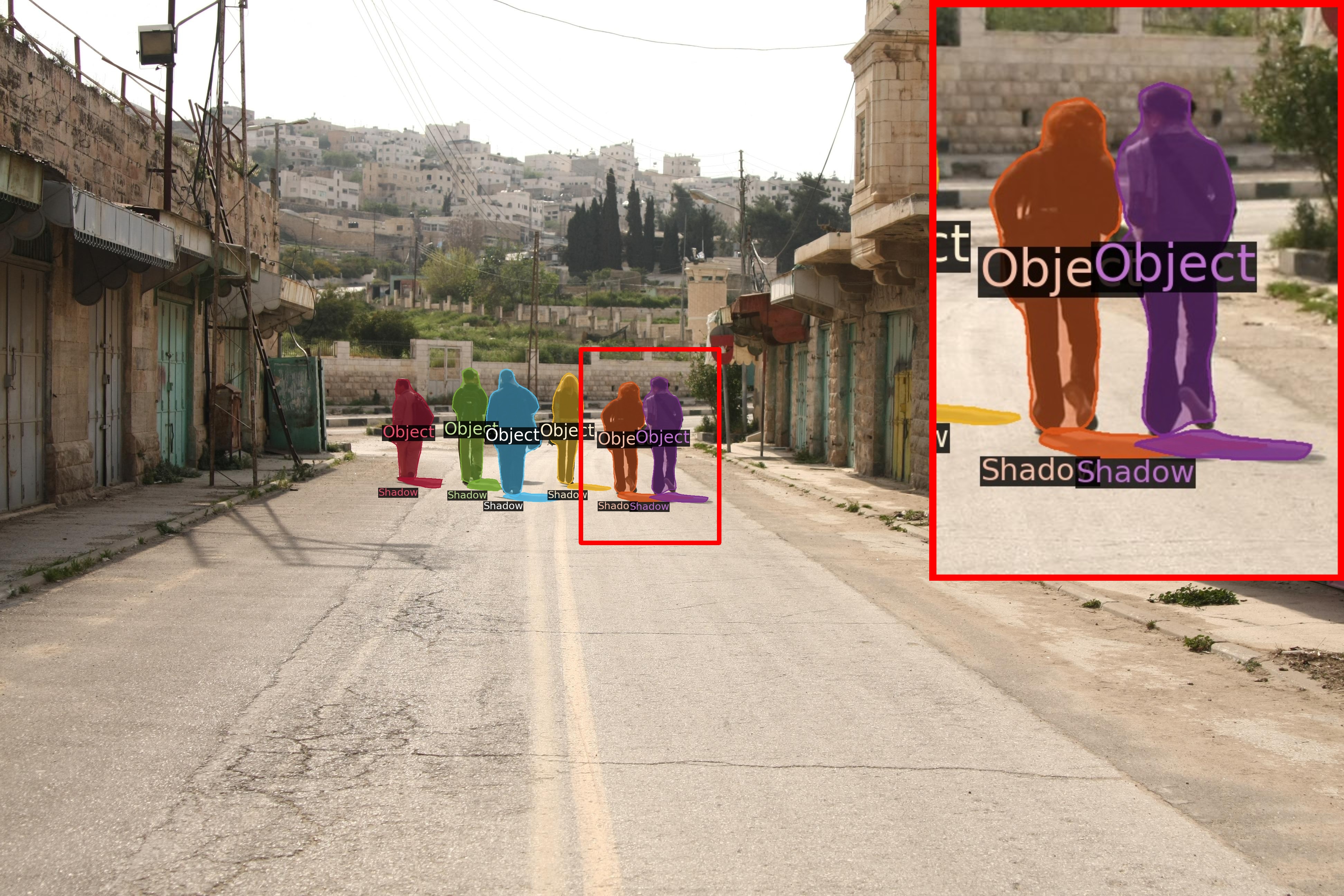}
        \end{minipage}
    \end{minipage}
    
    \vspace{0.8em}
    
    \begin{minipage}[c]{\linewidth}
        \begin{minipage}[c]{0.19\linewidth}
            \centering
            \includegraphics[width=\linewidth]{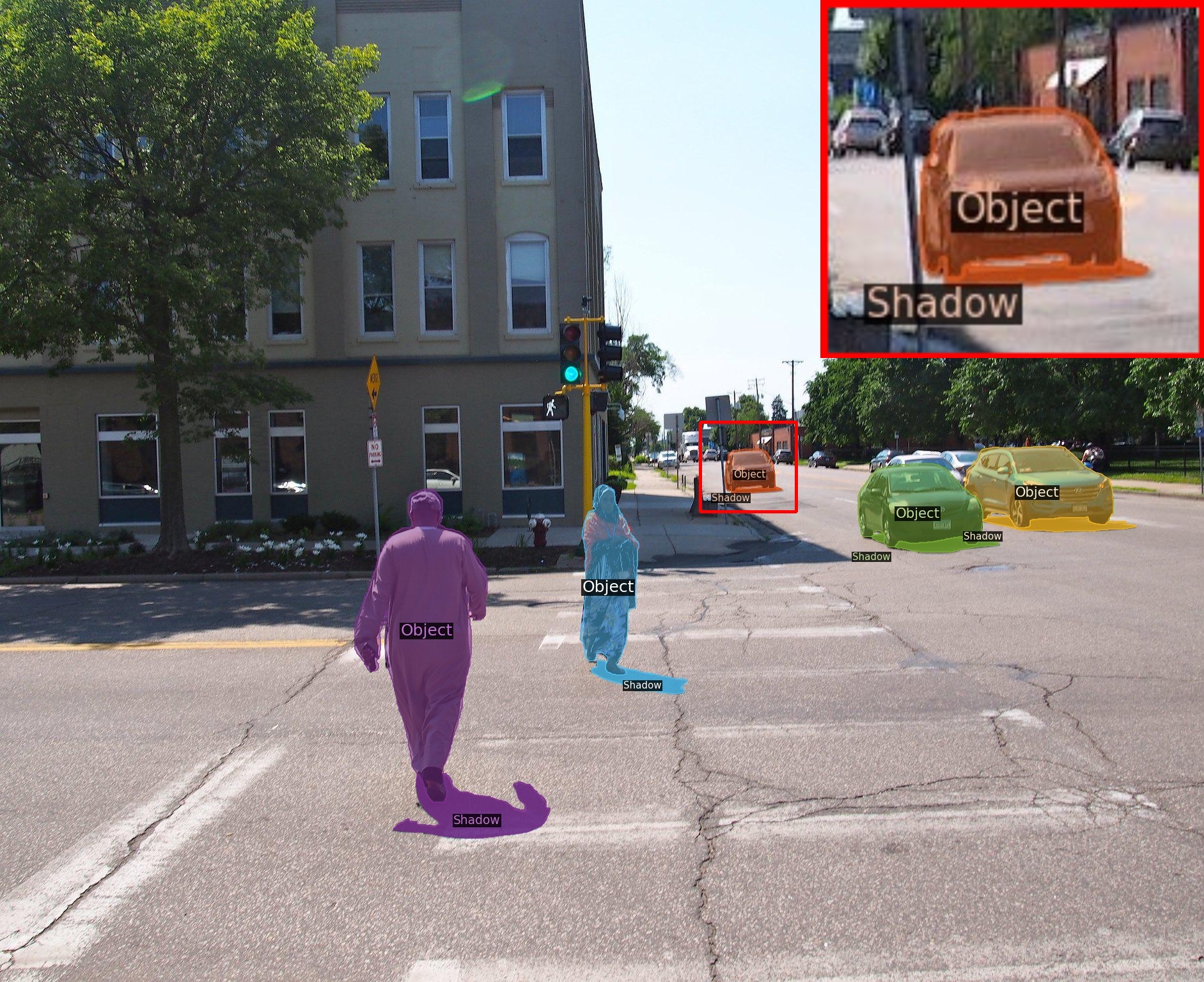}
            \subcaption*{(a) Input images}
        \end{minipage}
        \hfill
        \begin{minipage}[c]{0.19\linewidth}
            \centering
            \includegraphics[width=\linewidth]{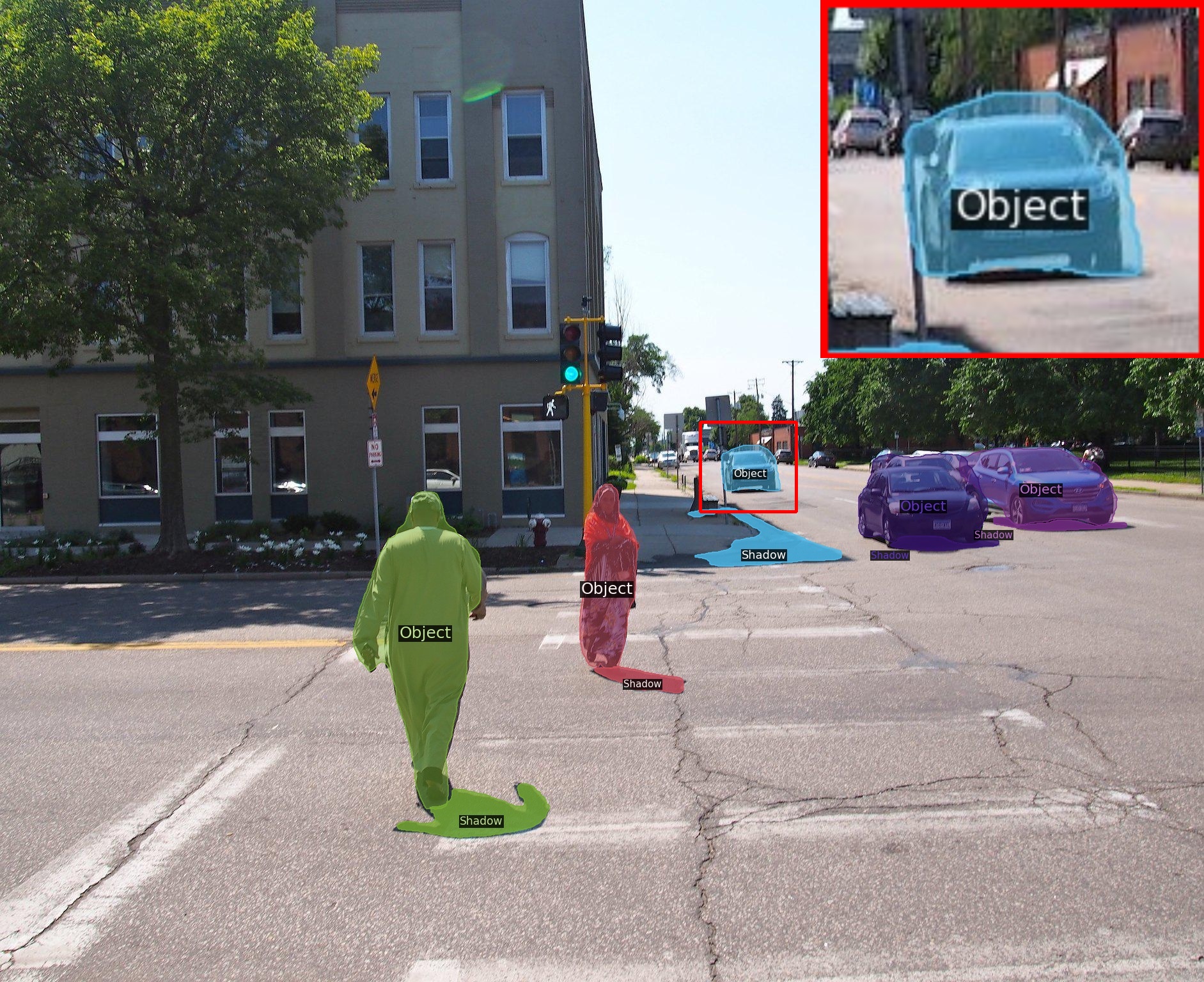}
            \subcaption*{(b) LISA~\cite{bib:lisa}}
        \end{minipage}
        \hfill
        \begin{minipage}[c]{0.19\linewidth}
            \centering
            \includegraphics[width=\linewidth]{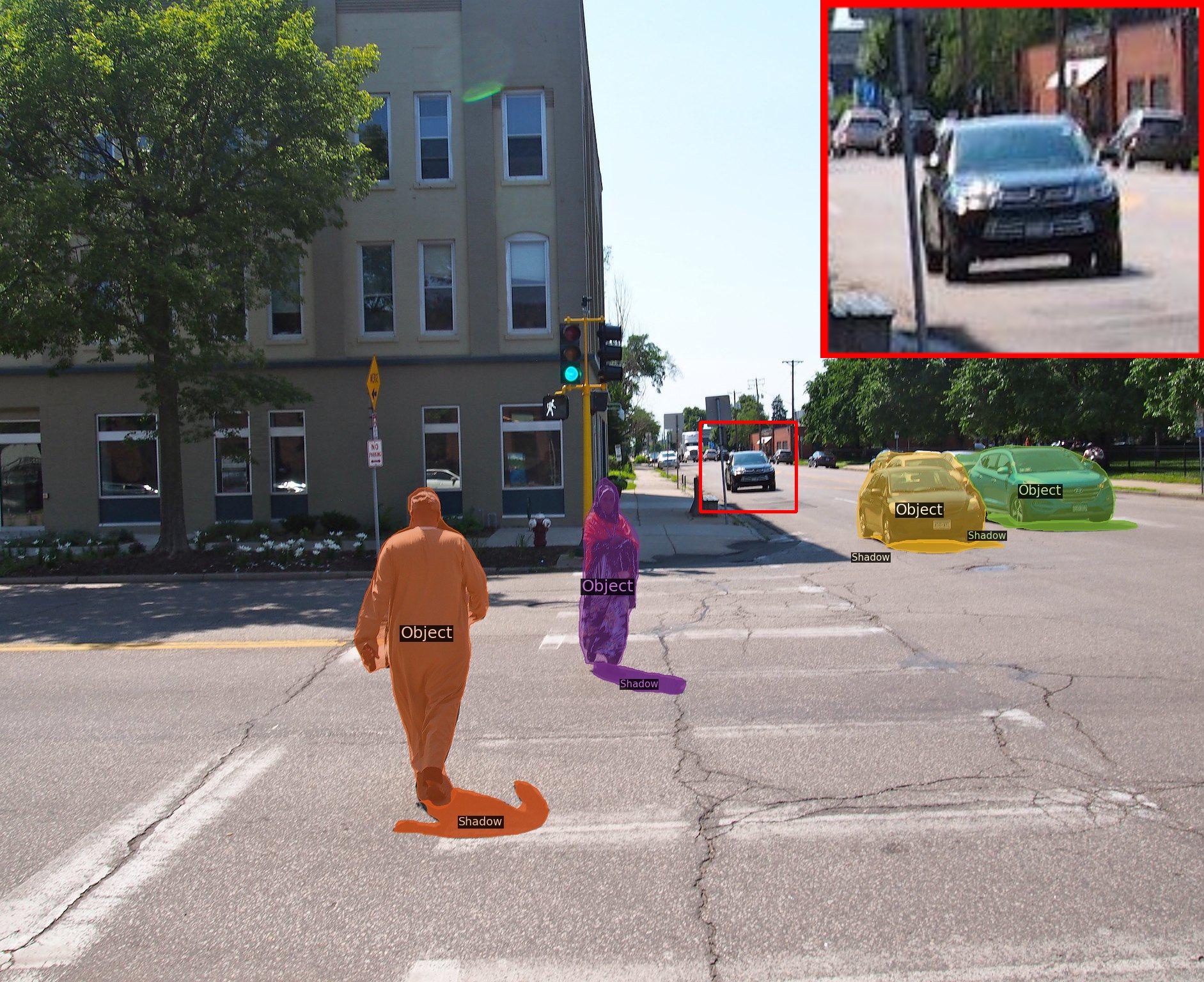}
            \subcaption*{(c) SSIS~\cite{bib:ssis}}
        \end{minipage}
        \hfill
        \begin{minipage}[c]{0.19\linewidth}
            \centering
            \includegraphics[width=\linewidth]{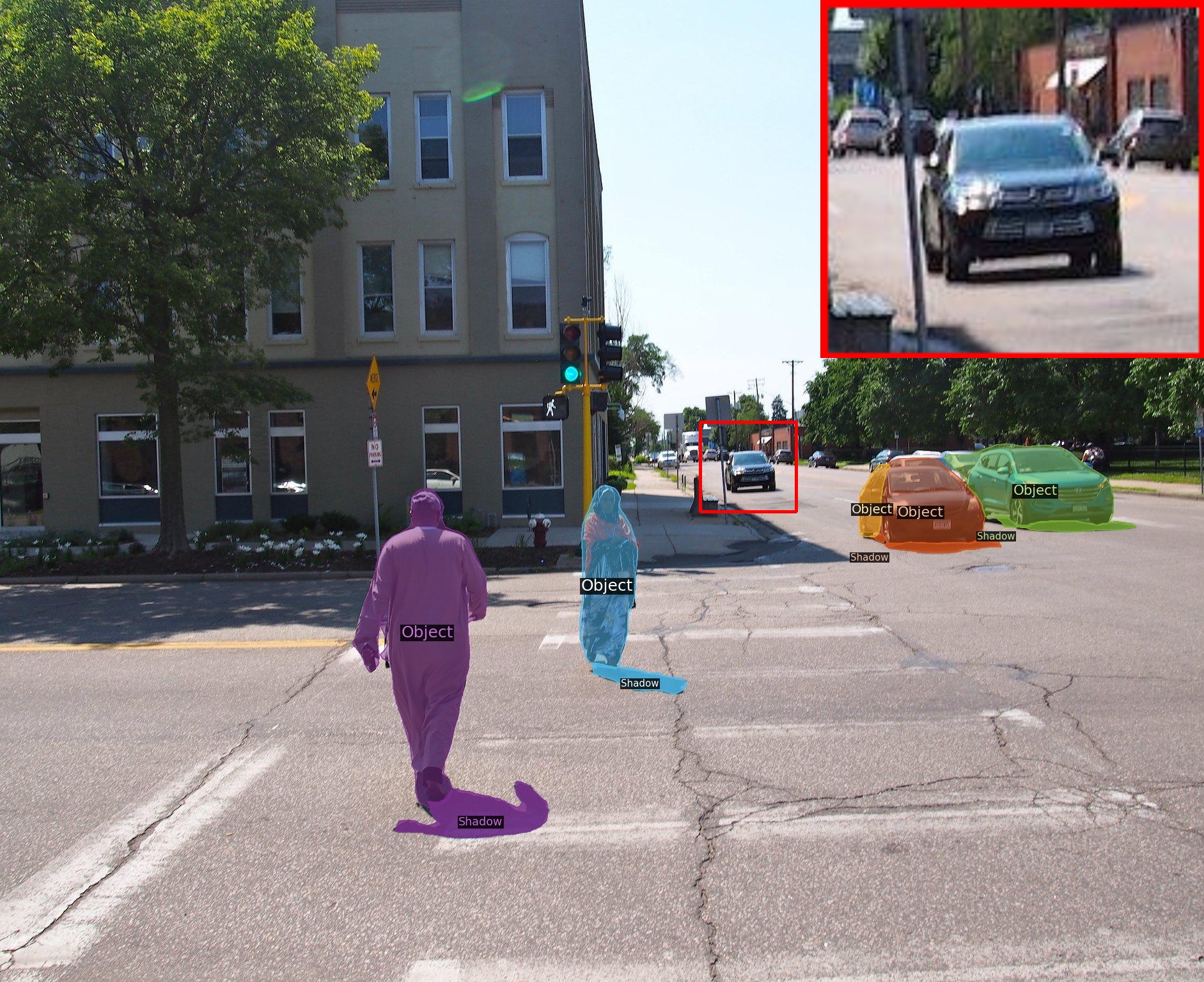}
            \subcaption*{(d) SSISv2~\cite{bib:ssisv2}}
        \end{minipage}
        \hfill
        \begin{minipage}[c]{0.19\linewidth}
            \centering
            \includegraphics[width=\linewidth]{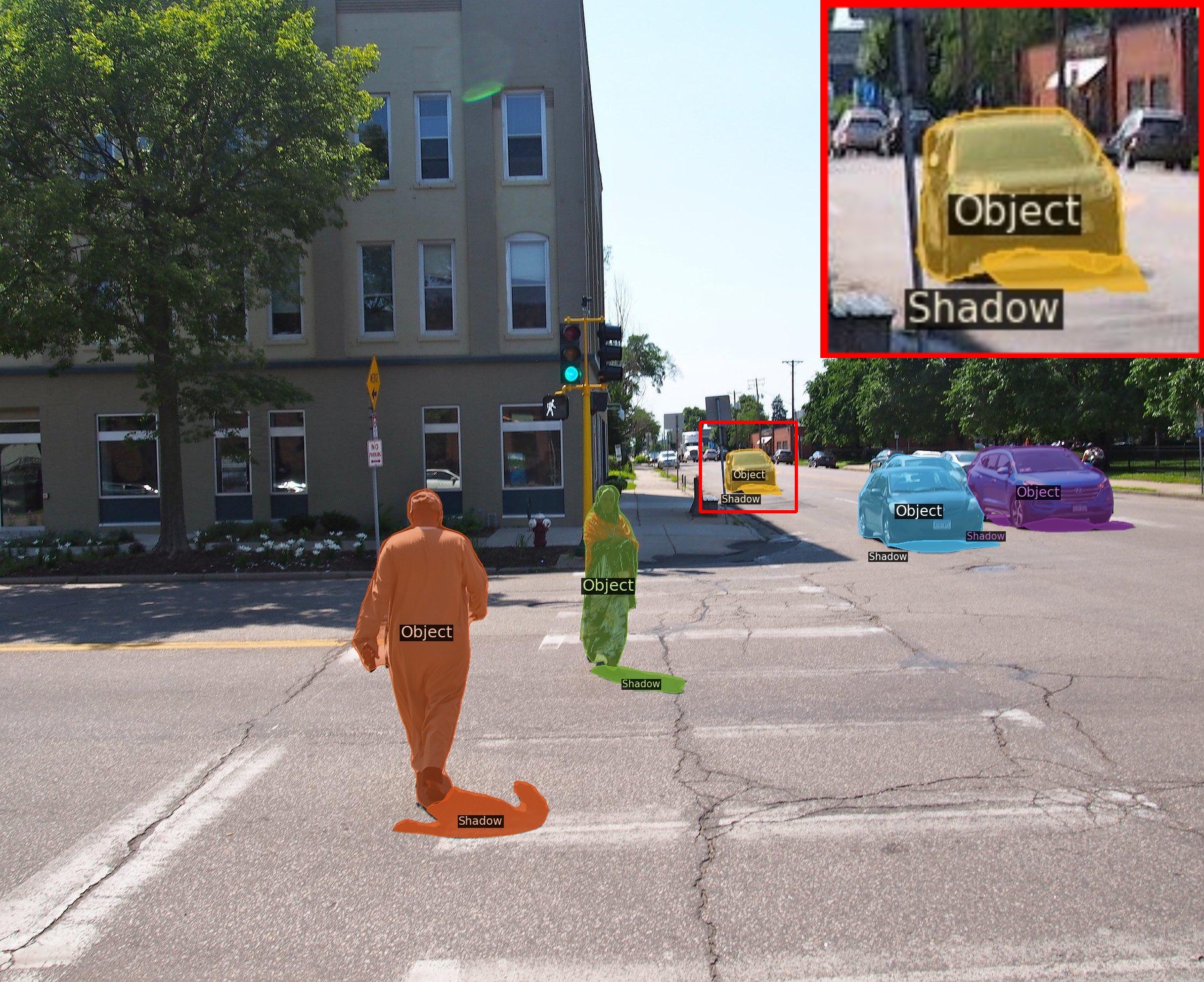}
            \subcaption*{(e) FIS-D3}
        \end{minipage}
    \end{minipage}
    \caption{\textbf{Visual comparison of some detections}.
    These results produced by various methods (b)-(e) on images (a) in the SOBA-testing or SOBA-challenge set.
    The regions in red boxes are zoomed for better visualization.}
    \label{fig:vis-comparison}
\end{figure*}

Next, we provide visual comparison results in \cref{fig:vis-comparison},
where (a) shows input images (b),(c),(d),(e) show the results produced using LISA~\cite{bib:lisa}, SSIS~\cite{bib:ssis}, SSISv2~\cite{bib:ssisv2} and FIS-D3, respectively.
The results show that
(i) FIS-D3 can detect shadows and objects more accurately, as shown in the first five rows and
(ii) FIS-D3 can discover more shadow-object association pairs, even for small objects, as shown in the last three rows.

\subsection{Additional analysis}
After introducing two training strategies, we then verified their effectiveness.
Additionally, we provide a more detailed analysis of their instance AP and fps.

\subsubsection*{Analysis of training strategies}
We evaluated the effectiveness of two proposed training strategies: shadow direction learning and box-aware mask loss.
Quantitative comparisons on SOBA-testing and SOBA-challenge are shown in \cref{tab:ap-soba-testing-strategy} and \ref{tab:ap-soba-challenge-strategy}, respectively.
The results demonstrate that integrating the shadow direction learning strategy improves all criteria.
In addition, the box-aware mask loss improves box-based detection metrics.
This reduces false positive detections that significantly deviate from ground truth regions,
obtaining appropriate AABB and improving box-based detection metrics.
Introducing both shadow direction learning and box-aware mask loss
result in the highest accuracy across all metrics on SOBA-testing and SOBA-callenge.
Consequently, it is demonstrated that our two strategies contribute to accuracy improvement.

\subsubsection*{Analysis of instance AP}
We conducted a detailed instance AP comparison between FIS and SSISv2, the current SOTA method.
We present the instance segmentation accuracy on the COCO~\cite{bib:coco} dataset for CondInst, which is the base model of SSISv2,
and FastInst, which is the base model of FIS, in the second and third rows of \cref{tab:inst-ap}.
This table shows that FastInst faces a limitation with low detection accuracy for small instances.
However, it is demonstrated that FIS-D3 is more accurate than SSISv2 for small instance detection on the SOBA dataset,
as shown in the last four rows of \cref{tab:inst-ap}.

\begin{table}[t]
    \centering
    \setlength{\tabcolsep}{4pt}
    \begin{tabular}{l|c|cccc}
        \hline
        method & dataset & AP & AP$_{\text{S}}$ & AP$_{\text{M}}$ & AP$_{\text{L}}$ \\
        \Xhline{2pt}
        CondInst & \multirow{2}{*}{\begin{tabular}{c} COCO \\ test-dev \end{tabular}} & 39.1 & \textbf{21.5} & 41.7 & 50.9 \\
        FastInst-D3 & & \textbf{40.5} & 11.9 & \textbf{57.9} & \textbf{76.7} \\
        \hline\hline
        SSISv2 & \multirow{2}{*}{\begin{tabular}{c} SOBA \\ -testing \end{tabular}} & 50.0 & 17.4 & 48.8 & 72.4 \\
        FIS-D3 & & \textbf{53.8} & \textbf{21.1} & \textbf{52.5} & \textbf{76.1} \\
        \hline\hline
        SSISv2 & \multirow{2}{*}{\begin{tabular}{c} SOBA \\ -challenge \end{tabular}} & 30.8 &  7.9 & 26.4 & 43.9 \\
        FIS-D3 & & \textbf{36.3} & \textbf{11.7} & \textbf{33.0} & \textbf{48.8} \\
        \hline
    \end{tabular}
    \caption{\textbf{Details of instance AP.}
    FastInst-D3~\cite{bib:fastinst} which is the base model of FIS-D3 demonstrated lower detection accuracy for small instances compared to CondInst~\cite{bib:condinst} which is the base model of SSISv2.
    However, when comparing FIS-D3 and SSISv2, FIS-D3 achieved higher detection accuracy for small instances.
    Since the original SSISv2 study~\cite{bib:ssisv2} did not report detailed instance AP,
    their results were reproduced using the official pretrained model from \url{https://github.com/stevewongv/SSIS}.}
    \label{tab:inst-ap}
\end{table}

\subsubsection*{Analysis of real-time performance}
The fps reported in \cref{tab:ap-soba-testing} fails to achieve real-time requirements ($\geq 30$ fps)
due to the large inference and input image sizes.
To properly assess FIS-D1's real-time performance,
this section evaluates the model using an appropriately sized inferences and input images.
The SOBA-testing is composed of images from the ADE20K~\cite{bib:ade20k-1, bib:ade20k-2},
SBU~\cite{bib:sbu-1, bib:sbu-2, bib:sbu-3}, ISTD~\cite{bib:istd}, and COCO~\cite{bib:coco} datasets, along with web-collected images.
Notably, images collected from the web in the SOBA-testing have higher resolutions,
which deviate from the standard image sizes of COCO test-dev, the benchmark for instance segmentation speed measurements.
When processing such large images, the time required to resize generated masks to their original resolutions becomes non-negligible.
Therefore, we validated FIS-D1's real-time performance on a dataset excluding web-collected images from SOBA-testing.
The dataset's average image size is $453 \times 584$ pixels, similar to COCO test-dev's average size of $484 \times 574$ pixels.
When inputting images to our model, we resized them to a shorter edge of 640 pixels and ensured a maximum length of 853 pixels.
The resulting inference speed and accuracy are presented in \cref{tab:realtime-fis}.
Here, FIS-D1 achieves real-time requirements and higher accuracy compared to SSISv2, the most accurate existing method.

\begin{table}[t]
    \centering
    \setlength{\tabcolsep}{4pt}
    \begin{tabular}{l|c|ccc}
        \hline
        \multirow{2}{*}{Network}
        & \multirow{2}{*}{fps}
        & \multirow{2}{*}{$SOAP_{segm}$}
        & \multirow{2}{*}{\begin{tabular}{c} Assoc. \\ $AP_{segm}$ \end{tabular}}
        & \multirow{2}{*}{\begin{tabular}{c} Inst. \\ $AP_{segm}$ \end{tabular}} \\
        &&&& \\
        \hline
        SSISv2 & 10.76 & 38.3 & 57.80 & 52.26 \\
        FIS-D1 & \textbf{32.21} & \textbf{39.3} & \textbf{60.57} & \textbf{52.27} \\
        \hline
    \end{tabular}
    \caption{\textbf{Instance shadow detection on a dataset excluding web-collected images from SOBA-testing.}
    Our FIS-D1 surpasses SOTA method in terms of mask-based detection metrics while enabling real-time processing.
    The result of SSISv2~\cite{bib:ssisv2} was reproduced using the official pretrained model.
    The input images of FIS-D1 were resized to a width of 640 pixels and a maximum length of 853 pixels,
    while the input images of SSISv2 were resized to a width of 800 pixels and a maximum length of 1,333 pixels.}
    \label{tab:realtime-fis}
\end{table}

\section{Conclusion}
This paper proposed a novel method for instance shadow detection
named FastInstShadow (FIS), which adopts a query-based method
to treat relationships between shadow and object features appropriately during the detection process
though existing methods detect shadows and objects independently.
The primary innovation of FIS is association transformer decoder,
composed of two dual-path transformer decoders: one to process object pixel
features and the other to process shadow pixel features.
In addition to the architectural modifications, shadow direction
learning and box-aware mask loss were designed to improve
instance shadow detection accuracy.
Our FIS has the capability to perform instance-shadow detection without the
pairing process of shadows and objects that has been indispensable to existing methods.

Experimental results using the SOBA datasets showed that detection accuracy and processing speed drastically improved:
FIS-D3 achieved the highest accuracy across all criteria, and FIS-D1 achieved the highest processing speed among all existing methods.
The processing speed of FIS-D1 reached more than 30 fps
for input images with moderate resolutions while maintaining better accuracy than SSISv2, the most accurate existing method.

\clearpage
{
    \small
    \bibliographystyle{ieeenat_fullname}
    \bibliography{defines,ref}
}

\end{document}